\documentclass[runningheads]{llncs}

\usepackage{eccv}

\usepackage{eccvabbrv}

\usepackage{graphicx}
\usepackage{booktabs}
\usepackage{pdfpages}
\usepackage{bbding}
\usepackage{float}
\usepackage{stfloats}
\usepackage{newunicodechar}
\newunicodechar{⁴}{\textsuperscript{4}}
\newunicodechar{⁵}{\textsuperscript{5}}

\usepackage[accsupp]{axessibility}  

\usepackage[pagebackref,breaklinks,colorlinks,citecolor=eccvblue]{hyperref}

\usepackage{orcidlink}

\begin{document}

\title{Quality Assured: Rethinking Annotation Strategies in Imaging AI} 


\author{Tim Rädsch\inst{1,2,3}\Envelope\orcidlink{0000-0003-3518-0315} \and
Annika Reinke\inst{1,2}\orcidlink{0000-0003-4363-1876}  \and
Vivienn Weru\inst{1}\orcidlink{0000-0002-7509-3307}  \and
Minu D. Tizabi\inst{1}\orcidlink{0000-0003-3687-6381} \and
Nicholas Heller\inst{4}\orcidlink{0000-0001-8516-8707}  \and
Fabian Isensee\inst{1,2}\orcidlink{0000-0002-3519-5886} \and
Annette Kopp-Schneider\inst{1}*\orcidlink{0000-0002-1810-0267} \and
Lena Maier-Hein\inst{1,2,3,5}*\Envelope \orcidlink{0000-0003-4910-9368}}

\authorrunning{T. Rädsch et al.}

\institute{ ¹German Cancer Research Center,	GER ²Helmholtz Imaging,	GER  ³Heidelberg University,	GER ⁴University of Minnesota, USA ⁵National Center for Tumor Diseases, GER  \hspace{0.7cm}   \Envelope:\{tim.raedsch; l.maier-hein\}@dkfz-heidelberg.de
            }

\maketitle

\begin{abstract}
This paper does not describe a novel method. Instead, it studies an essential foundation for reliable benchmarking and ultimately real-world application of AI-based image analysis: generating high-quality reference annotations. Previous research has focused on crowdsourcing as a means of outsourcing annotations. However, little attention has so far been given to annotation companies, specifically regarding their internal quality assurance (QA) processes. Therefore, our aim is to evaluate the influence of QA employed by annotation companies on annotation quality and devise methodologies for maximizing data annotation efficacy. Based on a total of 57,648 instance segmented images obtained from a total of 924 annotators and 34 QA workers from four annotation companies and Amazon Mechanical Turk (MTurk), we derived the following insights: (1) Annotation companies perform better both in terms of quantity and quality compared to the widely used platform MTurk. (2) Annotation companies' internal QA only provides marginal improvements, if any. However, improving labeling instructions instead of investing in QA can substantially boost annotation performance. (3) The benefit of internal QA depends on specific image characteristics. Our work could enable researchers to derive substantially more value from a fixed annotation budget and change the way annotation companies conduct internal QA. 


\begin{figure}
  \centering
  \includegraphics[width=1\textwidth]{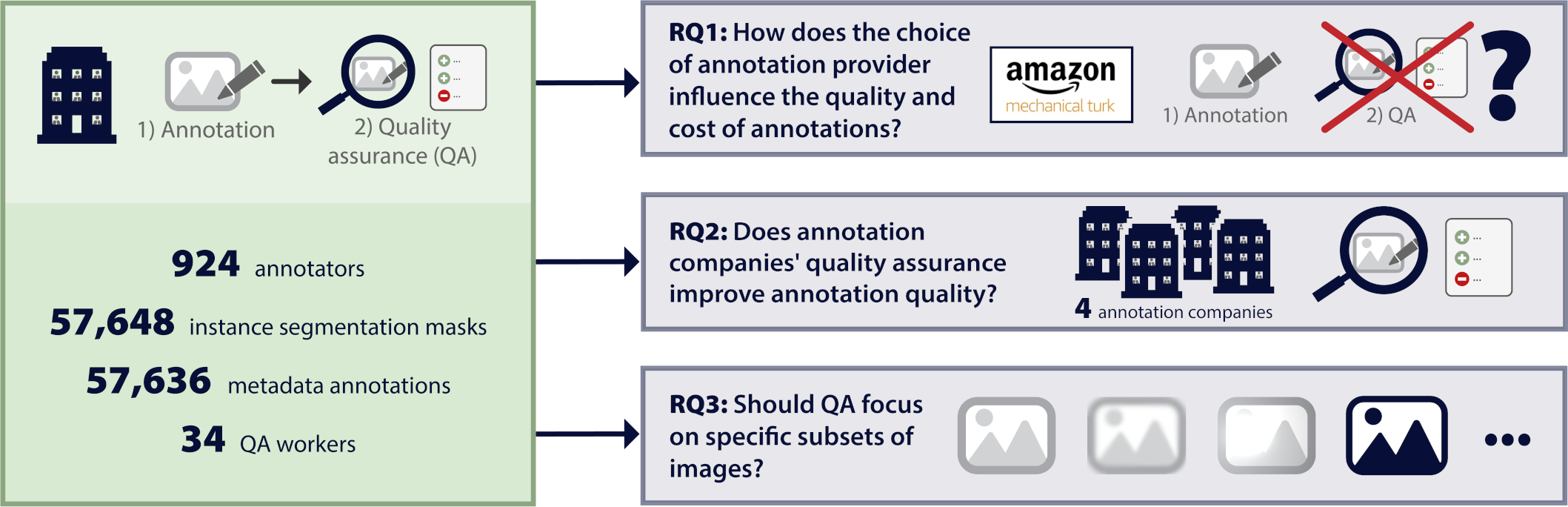}
  \caption{\textbf{Research Questions (RQs) tackled in this work.} Based on 57,648 instance segmentation masks annotated by 924 annotators and 34 quality assurance (QA) workers from five different annotation providers, we (1) compared the effectiveness of generating high-quality annotations between annotation companies and Amazon Mechanical Turk and (2) investigated the effects of annotation companies' internal QA and (3) real-world image characteristics on the annotation quality.}
  \label{fig:fig1}
\end{figure}

\end{abstract}

\section{Introduction}
\label{sec:intro}

Following a long period of extensive focus on pure method development, proper benchmarking and validation have recently been recognized as a major challenge in AI-based image analysis~\cite{sambasivan_everyone_2021, maier-hein_metrics_2024}. In this context, high-quality annotation of reference datasets is a fundamental requirement to ensure the meaningful validation of image analysis algorithms~\cite{halevy_unreasonable_2009}. As annotation tasks may be cumbersome, crowdsourcing is often used to partially outsource image annotation. Traditionally, such outsourcing is conducted via crowdsourcing platforms~\cite{maier-hein_can_2014}, with Amazon Mechanical Turk (MTurk)\footnote{https://www.mturk.com    
                      * shared last author}as the predominant choice~\cite{crequit_mapping_2018,orting_survey_2020}.

\begin{figure}
  \centering
  \includegraphics[width=1\textwidth]{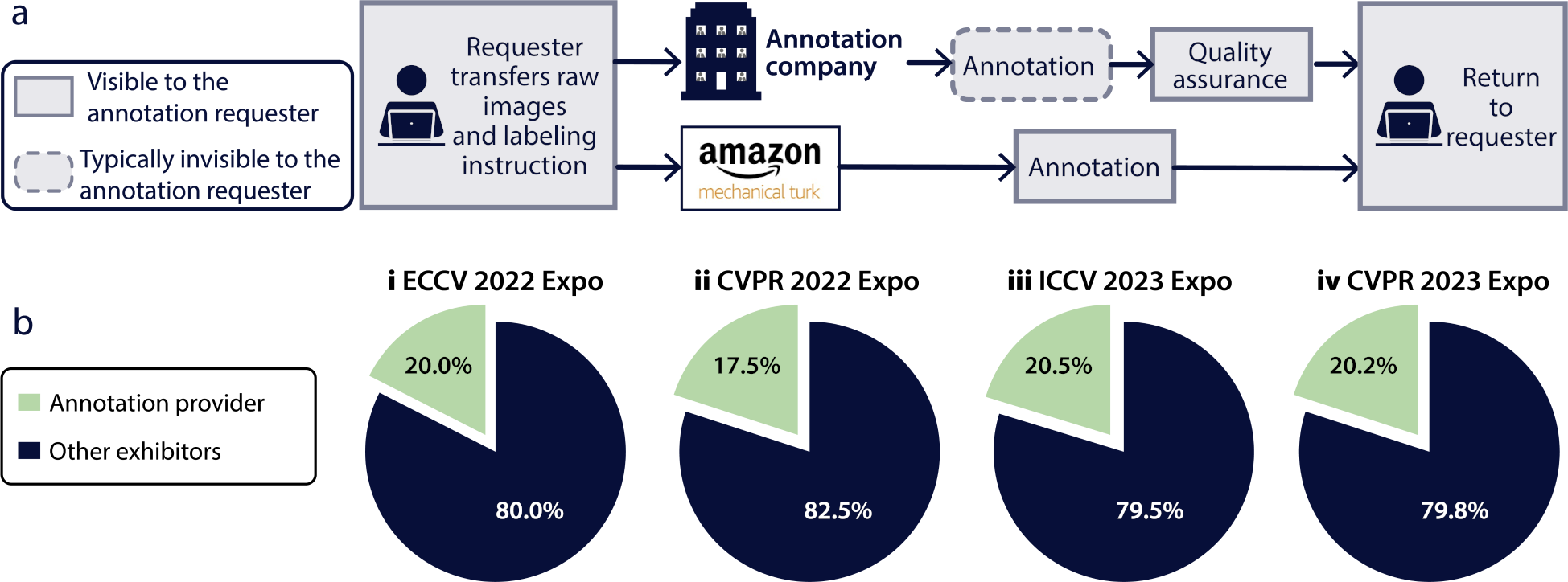}
  \caption{\textbf{a)} In contrast to Amazon Mechanical Turk (MTurk), annotation companies commonly perform quality assurance (QA) before delivering the result to the annotation requester.\textbf{ b) Annotation providers are an integral part of the computer vision community.} They account for around 20\% of exhibitors present at major computer vision conferences (ECCV 2022: n = 40, CVPR 2022: n = 97, ICCV 2023: n = 43, CVPR 2023: n = 104) in the last two years.}
  \label{fig:workflows_and_exhibitors}
\end{figure}

Recent literature has reflected the rise of annotation companies as a new type of provider with potentially better annotation performance compared to established crowdsourcing platforms such as MTurk ~\cite{graham_planetary_2022, le_ludec_problem_2023, radsch_labelling_2023, miceli_data-production_2022}. Annotation companies differ substantially from crowdsourcing platforms. First, they employ annotators directly and at shared office spaces~\cite{graham_planetary_2022,le_ludec_problem_2023}, enabling better external control of worker conditions and stricter data security measures, while crowdsourcing platforms can be accessed by anyone, without specific knowledge or training being required. Second, annotators in annotation companies typically work in small teams, with an experienced quality assurance (QA) worker being responsible for the team’s output. In the observed internal QA, QA workers make adjustments to the annotations generated by the team by reviewing each individual annotation and modifying those where they deem changes necessary. As a result, the annotation requester receives the quality-assured annotation, not the initial one (see Fig.~\ref{fig:workflows_and_exhibitors}\textcolor{red}{a}). In contrast, crowdsourcing platforms usually do not employ internal QA mechanics similar to annotation companies.

The importance of annotation providers for real-world applications is evident at the major computer vision conferences, where over the past two years, approximately 20\% of exhibitors were annotation providers (see Fig.~\ref{fig:workflows_and_exhibitors}\textcolor{red}{b}). While QA as a post-processing step for annotations from platforms such as MTurk has been researched (e.g.~\cite{daniel_quality_2019, le_ensuring_2010,heim_large-scale_2018,ipeirotis_quality_2010}), little is known about the benefit of internal QA on annotation quality within annotation companies. The only related work that we are aware of pertains to external QA from the perspective of an annotation company. Unlike in internal QA, in the case of external QA, domain experts or experienced annotators working on behalf of the requester check the annotation quality as a post-processing step of the received annotations~\cite{zheng_improving_2023, lu_rethinking_2023}. Our paper, on the other hand, focuses on the internal QA processes of an annotation company. 

In light of the vast sums of money to be spent on QA~\cite{report_annotation_2030}, the purpose of this paper is to shed light on whether annotation companies' internal QA benefits the annotation quality. Based on more than 50,000 instance segmentation masks annotated by more than 900 annotators from five different annotation providers, we investigate the following research questions (RQs; see Fig.~\ref{fig:fig1}):

\textbf{RQ 1) How does the choice of annotation provider influence the quality of annotations?} As the selection of the annotation provider is among the first and key decisions in any annotation project after data collection and generation of labeling instructions, we first investigated companies' effectiveness in generating annotations, which is particularly relevant in cases of constrained annotation budgets requiring a cost-effective strategy. As both quantity and quality of annotated images for a given budget are relevant in provider selection, we compared the annotation quality and costs of five different providers, including MTurk and four annotation companies, on identical images.

\textbf{RQ 2) Does annotation companies' quality assurance improve annotation quality?} A major decision in the design of an annotation process concerns the allocation of resources. A common perception is that internal QA of annotation companies, following the initial annotation, substantially enhances the quality. Here, we challenged this assumption by investigating the improvement in annotation quality from internal QA compared to that achieved through enhanced labeling instructions.

\textbf{RQ 3) Should quality assurance focus on specific subsets of images?} In theory, the efficacy of QA may be contingent upon specific image characteristics~\cite{joskowicz_inter-observer_2019,ros_how_2021,idrissi_imagenet-x_2022}. For instance, underexposed images might benefit more from a secondary review compared to those with optimal illumination. Concurrently, images that present annotation challenges for human observers might be particularly pertinent for the training and validation of algorithms~\cite{ros_how_2021,idrissi_imagenet-x_2022}. Here, we investigated whether and to what extent the benefit of QA depends on specific image characteristics. 


\section{Related Work}
\label{sec:related_work}

\subsection{Comparison of annotation providers}
\label{ssec:rw-annotation-providers}
With respect to the quality of annotations obtained from varying annotation providers, the closest work conducted are comparisons between crowdsourcing platforms and domain experts. Irshad et al.~\cite{irshad_crowdsourcing_2014} and Petrović et al.~\cite{petrovic_crowdsourcing_2020}, among others, have shown that crowdsourced annotations can reach the annotation quality of domain experts in safety-critical applications. In recent literature, Duggan et al.~\cite{duggan_gamified_2023} and Kentley et al.~\cite{kentley_agreement_2023} compared domain-specialized crowds as a new type of annotation provider and found comparable performance to domain experts. Furthermore, Rädsch et al.~\cite{radsch_labelling_2023} provided a comparison of annotation companies to crowdsourcing, but focused solely on raw annotation quality without QA and without relating it to annotation budgets.

Interestingly, researchers report concerns about the annotation quality obtained from MTurk~\cite{lu_rethinking_2023}, such as obtaining unusable annotations in up to 67.8\% of cases~\cite{cheplygina_early_2016}. Other research fields, such as psychology, even report a \textit{"substantial decrease in data quality"}~\cite{chmielewski_mturk_2020}. However, we are not aware of work that has systematically compared the performance between annotation companies and crowdsourcing platforms such as MTurk while considering annotation budgets.
\subsection{Quality assurance}


The main goal of QA is to enhance and ensure the quality of annotations. A comprehensive overview on QA in crowdsourcing platforms is provided by Daniel et al.~\cite{daniel_quality_2019}. However, their research does not include annotation companies. During the annotation process, quality control techniques generally fall into three categories~\cite{lu_rethinking_2023}: (1) comparing to a gold standard with pre-defined reference annotations (e.g.,~\cite{le_ensuring_2010}), (2) analyzing combined outputs from multiple annotators (e.g.,~\cite{ipeirotis_quality_2010}),  and (3) investigating annotator behaviors (e.g.,~\cite{heim_large-scale_2018,hutson_quality_2019}). Other approaches to enhancing annotation quality on crowdsourcing platforms target training of crowdworkers~\cite{dow_shepherding_2012} and iterative refinement of the annotation environment~\cite{gaikwad2017prototype,bragg_sprout_2018, chang_revolt_2017}. It should be noted that annotation byproducts, such as mouse movement during annotation generation~\cite{han2023neglected,heim_large-scale_2018} or inter-rater agreement~\cite{yu_difficulty-aware_2020}, can be used to boost model performance, by incorporating this information in the training.

The closest related work focusing on the impact of QA in annotation companies was introduced by Wang et al.~\cite{zheng_improving_2023} and Lu et al.~\cite{lu_rethinking_2023}. However, their work focuses on external QA conducted by experienced personnel from the requester rather than internal QA, and features a substantially smaller number of annotators from annotation companies. To our knowledge, a systematic investigation into the effects of internal QA processes within annotation companies is thus, so far, lacking.
\label{ssec:rw-qa}

\subsection{Impact of real-world image characteristics on annotation quality}
\label{ssec:rw-image-characteristics}
Related research analyzing the impact of real-world image characteristics highlights their effect on annotations created by machine learning models and annotators alike. Michaelis et al.~\cite{michaelis_benchmarking_2020} demonstrated that models easily encounter failures with increasingly challenging real-world image characteristics, such as added snow or poor lighting characteristics. Similarly, annotators generate less reliable annotations in the presence of challenging real-world image characteristics. This effect can be observed both within~\cite{joskowicz_inter-observer_2019, lu_rethinking_2023, ali_objective_2020} and across domains~\cite{hettiachchi_challenge_2021,kohli_medical_2017}. However, a systematic analysis of the impact of such characteristics on internal QA employed by annotation companies is lacking.
\section{Methods}
\label{sec:Methods}

In the following, we describe the selected data, the labeling instructions, annotation providers and annotators, the QA process, and the experimental design of our work.

\subsection{Data}
\label{ssec:methods-data}
We conducted our experiments on the instance segmentation data from the Heidelberg Colorectal (HeiCo) dataset for surgical data science in the sensor operating room~\cite{maier-hein_heidelberg_2021}, representing a safety-critical application that crowdsourcing platforms can still handle. The task involved segmenting surgical instruments from laparoscopic videos, resulting in 57,648 instance segmentation masks on 4,050 unique frames. Furthermore, we used 57,636 metadata annotations on the unique images to quantify the effect of varying real-world image characteristics (see RQ3). Characteristics included motion blur, underexposure, object occlusion, or overlapping objects, among others, and are illustrated in Suppl.~\ref{sec:app_A_image_characteristics}.

\subsection{Labeling instructions}
\label{ssec:methods-li}
To investigate the benefit of QA compared to an investment in detailed labeling instructions, we created three distinct types of labeling instructions with varying levels of information density, namely, (1) minimal text, (2) extended text, and (3) extended text including exemplary pictures. Each subsequent labeling instruction type was structured to contain more detailed information than the preceding. To ensure accurate labeling of the imaging data, we established a collection of design principles applicable to all labeling instructions:
\\
(1) Labeling information is presented in a slide format to enhance the processing of information by humans.
\\
(2) Each slide contains a distinct 'chunk' of information, which is a self-contained piece of content. This approach of 'chunking' information lessens the cognitive load on the working memory.
\\
(3) Related information chunks are placed in close proximity to one another.
\\
(4) A uniform design template is used throughout, featuring specific fonts, symbols, and color schemes.\\
The three labeling instructions differ as follows:

Minimal text labeling instructions:
These instructions provide a concise textual outline, offering examples of typical annotations and highlighting the most frequent annotation types (see Suppl.~\ref{sec_sub:li_min}). They reflect a scenario with minimal effort invested in developing the instructions. The basic framework includes seven slides totaling 168 words.

Extended text labeling instructions:
Building upon the simplified version, the extended text instructions offer a thorough textual explanation, supplemented by both positive and negative examples (refer to Suppl.~\ref{sec_sub:li_ext}). They cover both regular and some rare annotation scenarios. The enhanced version comprises ten slides with a total of 446 words.

Extended text including pictures labeling instructions:
These instructions augment the enhanced textual guidelines with images (detailed in Suppl.~\ref{sec_sub:li_pic}). The visual aids feature descriptions, symbols, annotations, and color coding to effectively communicate the details on the slides. They also include illustrations of rare annotation cases, indicating a deep and well-documented understanding of the labeling process. This set contains 16 slides with a total of 961 words. We refer the reader to Suppl.~\ref{sec:app_B_li} for full details of each labeling instruction type.

\subsection{Annotation providers}
\label{ssec:methods-annotation-providers}
The study was conducted based on 216 annotators and 34 QA workers from four annotation companies and 708 crowdworkers from the crowdsourcing platform MTurk. Our goal was to provide the best representation for both annotation provider types. 

Consequently, we established higher standards for our MTurk crowdworkers compared to the usual research benchmarks, which generally require a 95\% acceptance rate of Human Intelligence Tasks (HITs) and at least 100 completed tasks~\cite{cheplygina_early_2016, heim_large-scale_2018, bragg_sprout_2018}. Our criteria included a 98\% acceptance rate with over 5,000 completed HITs. We spread out the HITs over 40 days at all times to obtain a representative MTurk sample and compensated workers fairly, as recommended by Litman et al.~\cite{litman_relationship_2015}. The annotation companies we selected have a solid background in handling large-scale annotation tasks in a variety of domains and operate internationally. The annotators were allocated by the companies themselves to replicate the typical workflow of a project carried out by an annotation company. Each annotator was linked to a specific annotation provider that had been predetermined. No selection process from our side took place for either the annotators, QA workers, or the MTurk crowdworkers.  

To prevent information leakage, we conducted the labeling process with increasingly detailed labeling instructions (minimal text instruction, extended text instruction, and extended text including pictures instruction). Each annotator worked exclusively with one of the three instruction types as introduced in Sec.~\ref{ssec:methods-li}. To enhance security, we included a mandatory minimum gap of ten days between consecutive labeling instructions.

\subsection{Quality assurance process}
\label{ssec:methods-qa}
In the internal QA process, annotators typically report to a small number of experienced QA workers, who are often responsible for training their team and, in our case, responsible for the QA of the annotations.

The general workflow for all companies consisted of an annotation stage, in which images were assigned randomly to each annotator, and a QA stage. During the annotation stage, annotators received raw images with the task of annotating them until they considered the quality of annotations satisfactory. Annotators were aware that their work would be checked by QA. Once submitted, the QA worker gained access to the generated annotations, allowing them to make modifications until they were satisfied before submitting the final annotations for delivery to the requester. Throughout this process, both annotators and QA workers retained access to the respective labeling instructions at any given moment. We obtained access to the annotations both before and after QA through a software company handling a substantial number of annotation projects annually. The experiments were seamlessly integrated into the software company’s pipeline of annotation projects, reducing the risk of exposing our experiments. Pricing was established to be representative of other projects of similar size.

To maintain fairness and avoid giving any annotation provider a potential informational advantage that may have affected our statistical analysis, no specific queries from the annotation companies about the content of the labeling instructions were addressed.

\subsection{Experimental design}
\label{ssec:methods-exp}
\textbf{RQ 1:} Each of the five labeling providers annotated the same images with each set of labeling instructions, employing separate annotators for each task. We analyzed the number of spam images, images with severe errors, i.e., images with at least one false positive and/or false negative, and the overall annotation quality with the Dice Similarity Coefficient (DSC)~\cite{dice_measures_1945} as the overlap-based metric and the Normalized Surface Distance (NSD)~\cite{reinke_understanding_2024} as the distance-based metric.
We define a spam annotation as a purposefully poor-quality annotation with the intent to obtain the payment without properly conducting the annotation task. Annotation duration was used as an auxiliary measure, where accessible. For instance, if an annotator completed the task in only 17 seconds, they could not have read the instructions and properly performed annotations on several images. Of note, a spammer could wait a relevant number of seconds and then enter a poor-quality annotation to bypass the annotation duration check.
Thus, each image was additionally visually inspected by an engineer with extensive labeling experience. Spam annotation examples are displayed in Suppl.~\ref{sec:app_D_li}.

\textbf{RQ 2:} The annotation companies conducted their internal QA processes directly after their annotators completed the annotation job, as explained in Sec.~\ref{ssec:methods-qa} and Fig.~\ref{fig:workflows_and_exhibitors}\textcolor{red}{a}. The experiments began with only providing the minimal text labeling instructions, followed by the extended text labeling instructions, and finally, the extended text instructions with pictures, to prevent information leakage. 

For each annotated image, we calculated the DSC and NSD per instance, aggregated the scores per annotated image, and aggregated the scores over annotation company and labeling instruction. We then compared the annotation quality without and with QA across the three types of labeling instructions for the four annotation companies by computing both absolute and relative changes in DSC/NSD aggregates, the percentage of modified images, and the change in the percentage of severe errors, among other parameters.

\textbf{RQ 3:} In contrast to MTurk, the four annotation companies conducted an additional annotation round, including QA, with the extended text including pictures labeling instructions to quantify the impact of different real-world image characteristics, represented by 57,636 metadata annotations,  on the probability of improvement. The data was analyzed with a logistic mixed model~\cite{mcculloch_generalized_2004} which included random effects for the image, the annotator, and the annotation company.	
Further method details are provided in Suppl.~\ref{sec:app_C_li}.

\begin{figure}[tb]
  \centering
  \includegraphics[width=0.8\textwidth]{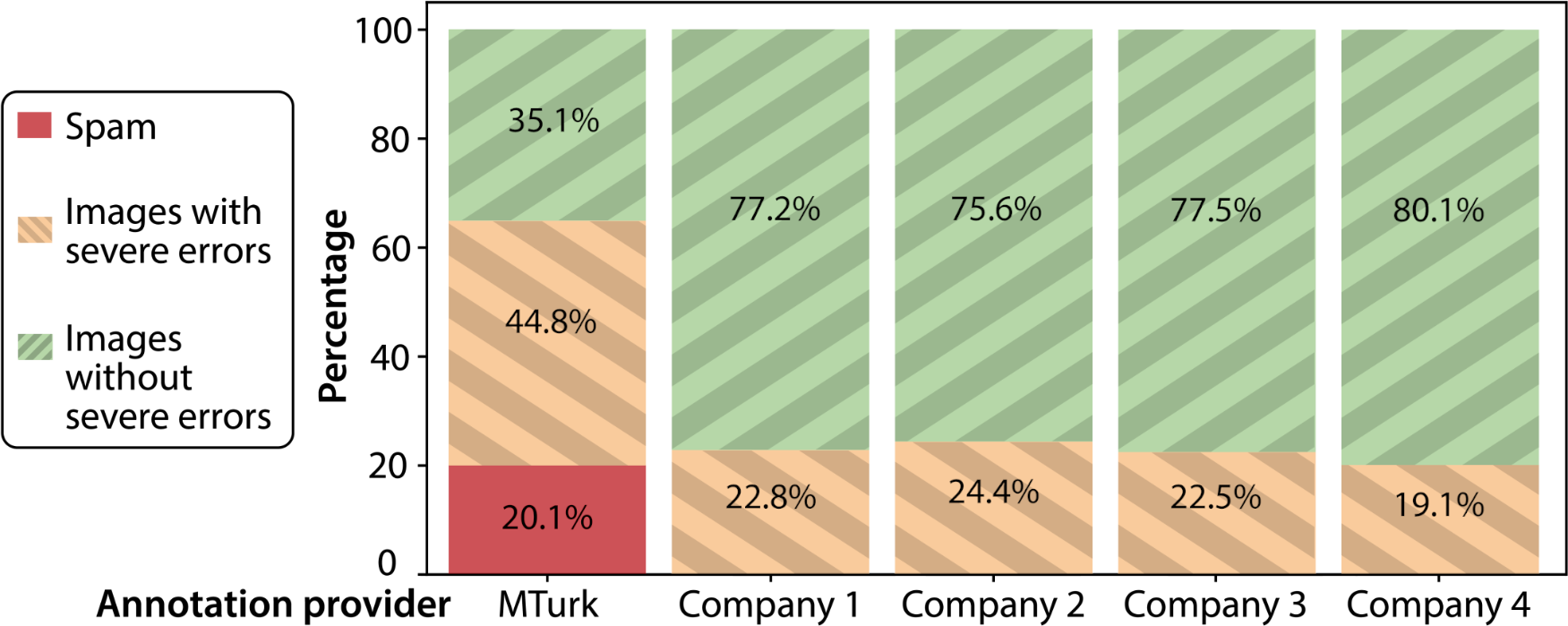}
    \caption{\textbf{Annotation companies are more efficient at generating high-quality images than Amazon Mechanical Turk (MTurk).} Percentage of images without severe errors (green), images with severe errors (orange), and spam (red) obtained for the five different annotation providers investigated. Annotation costs per image were lower for all companies compared to MTurk (median: 61.1\%).}
  \label{fig:budgets}
\end{figure}

\section{Results}
\label{sec:results}

\subsection{How does the choice of annotation provider influence the quality and cost of annotations?}
\label{ssec:results-rq1}
Annotated images generated by the annotation companies were of higher quality, comprising no spam (MTurk: $\sim$20\%) and a much higher proportion of remaining images without severe annotation errors (factor of $>$ 2 improvement for all companies compared to MTurk), as shown in Fig.~\ref{fig:budgets}. In addition, the annotation companies were 61.1\% cheaper in the median, despite their additional employment of QA workers. More specifically, the costs ranged from 24.5\% to 81.5\% of the MTurk costs for the same images and number of annotations. We have not been given permission to reveal the identities and exact absolute budgets used for  the annotation companies, hence we report costs relative to the MTurk budget. Examples of spam annotations are provided in Suppl.~\ref{sec:app_D_li}.

Regardless of the annotation scenario, the annotation companies generated higher quality annotations than MTurk with higher DSC scores including smaller interquartile ranges (IQRs) (see Fig.~\ref{fig:provider_comparison_dsc}),  higher NSD scores including smaller IQRs (see Fig.~\ref{fig:provider_comparison_nsd}), and lower numbers of severe errors (see Fig.~\ref{fig:provider_comparison_sev_errors}).

\begin{figure}[t]
  \centering
  \includegraphics[width=1\textwidth]{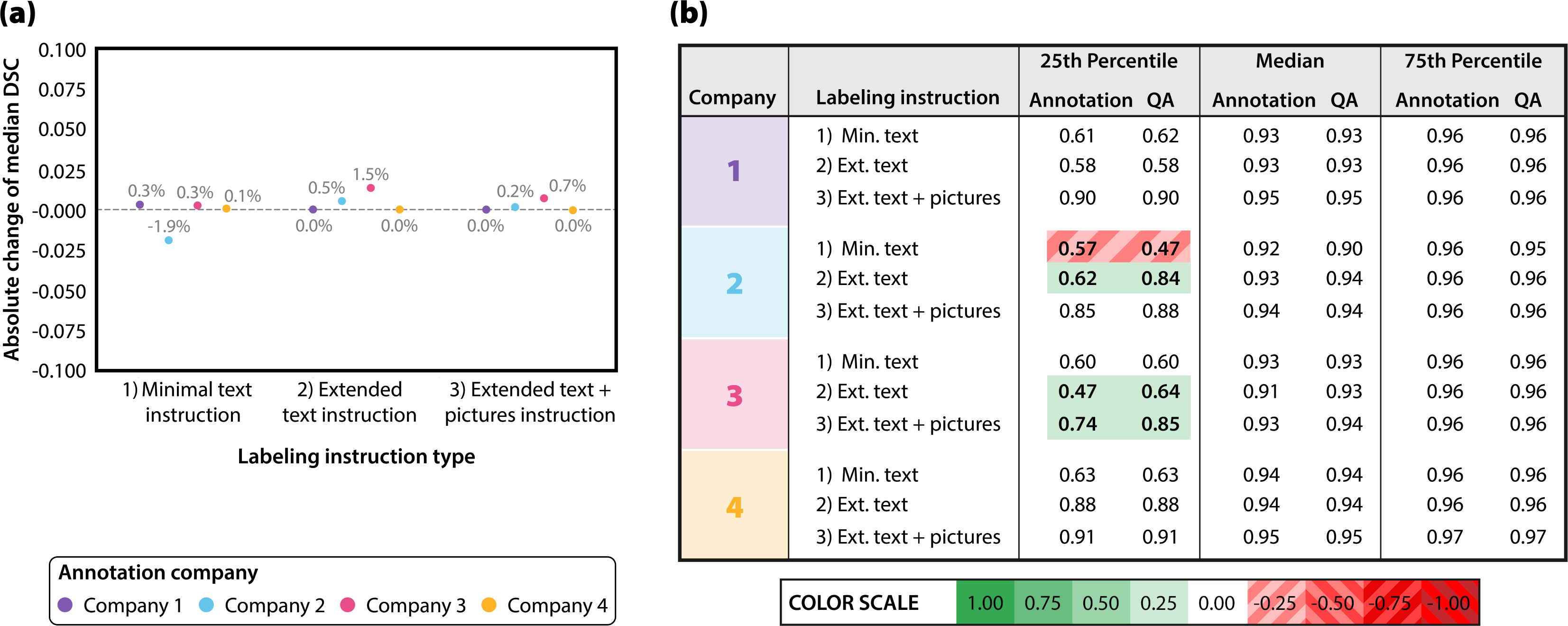}
  \caption{\textbf{QA of annotation companies only provides marginal improvements, if any. These depend highly on the company and type of labeling instruction. a:} Absolute change of the median DSC resulting from performing QA. For each company and labeling instruction, the median was obtained by aggregating the DSC scores over the images. \textbf{b:} Corresponding absolute values of the DSC. Positive and negative changes from annotation to QA are colored according to the color scale.}
  \label{fig:changes}
\end{figure}

\begin{figure}[t]
  \centering
  \includegraphics[width=1\textwidth]{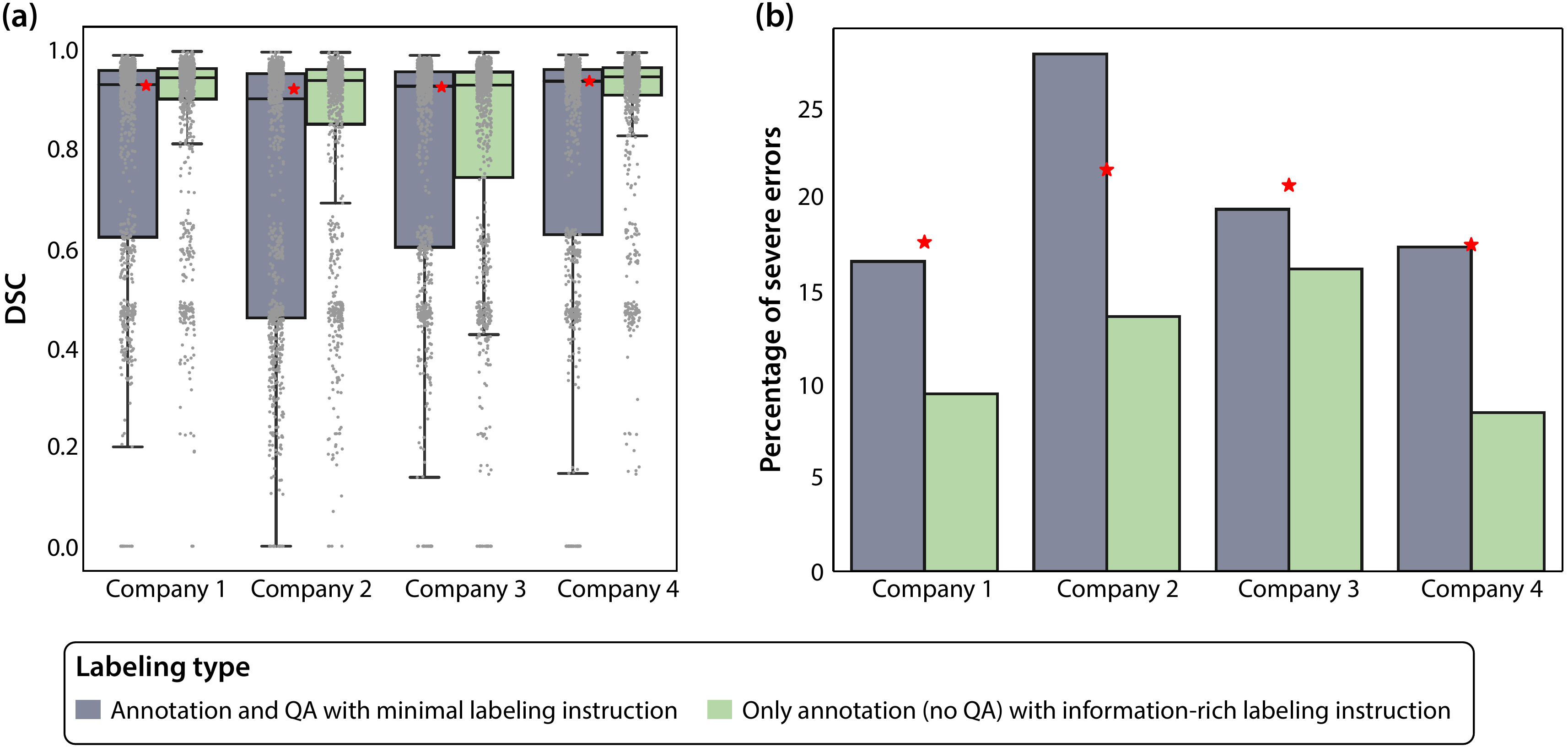}
  \caption{\textbf{Improving the labeling instructions yields a higher effect compared to adding QA to the annotation process.} The baseline performance (minimal labeling instruction, no QA) according to the DSC (a) and the percentage of severe errors (b) is shown individually per company with a red asterisk (minimal labeling instruction, no QA). While adding QA only yields marginal improvements (dark blue), major performance boosts are obtained by improving the labeling instructions provided to the annotators (green). DSC scores were aggregated for each image and per labeling type. DSC displayed as dots and box plot (band indicates the median, box indicates the first (25th percentile) and third (75th percentile) quartiles). \(0 \leq \text{DSC} \leq 1\).}
  \label{fig:boxplot}
\end{figure}

\subsection{Does annotation companies' quality assurance improve annotation quality?}
\label{ssec:results-rq2}
As shown in Fig.~\ref{fig:changes}, the internal QA conducted by annotation companies does not substantially improve annotation quality if the QA workers only use minimal text instructions. The impact of QA when having access to extended text instructions or extended text including pictures instructions depended highly on the annotation company, while the effect was still minimal compared to not performing QA: Two of the four annotation companies show no change in the median DSC when comparing QA to non-QA outputs for extended text or extended text and picture instructions. The remaining two companies only show a very small relative improvement of below 2\%. 

We obtained even smaller improvements with the NSD scores, as depicted in Suppl. \textcolor{red}{E}. In certain scenarios, QA was able to remove a small proportion of severe errors, however, the vast majority remains (see Suppl. \textcolor{red}{F}). Notably, the percentage of modified images by QA workers depended highly on the company, as depicted in Suppl. \textcolor{red}{G}. Fig.~\ref{fig:boxplot} further illustrates that providing annotators with high-quality labeling instructions including pictures (green) leads them to vastly outperform annotators working with lower-quality labeling instructions, even if these are followed by additional internal QA (dark blue).

\subsection{Should quality assurance focus on specific subsets of images?}
\label{ssec:results-rq3}
\begin{figure}[t]
  \centering
  \includegraphics[width=1\textwidth]{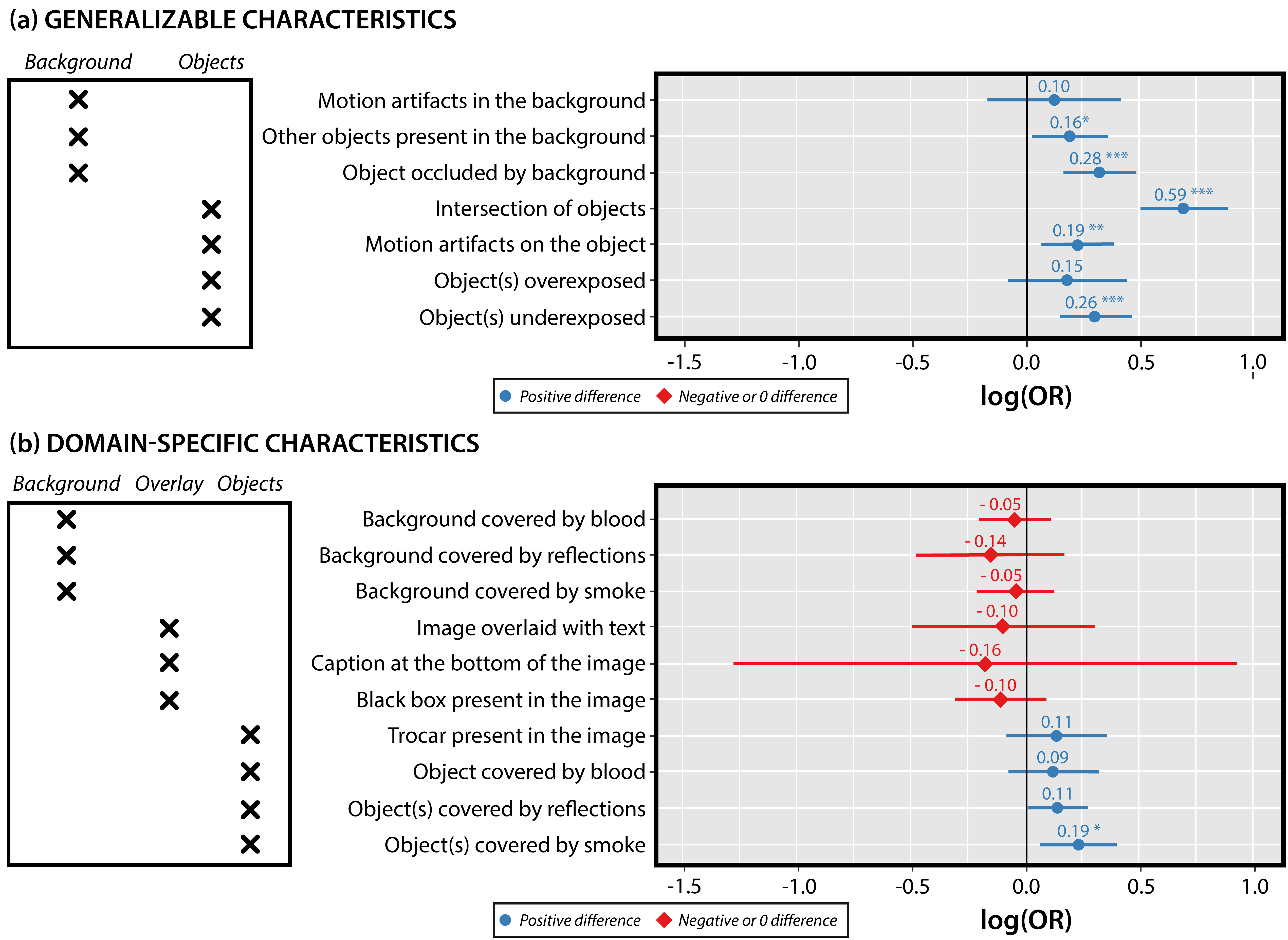}
  \caption{\textbf{QA of annotation companies can be beneficial to certain images with difficult real-world generalizable image characteristics, such as underexposure or intersection of objects.} We display the impact of (a) real-world generalizable and (b) domain-specific image characteristics on the odds of improvement. Improvement is defined as a positive difference in the QA, while no improvement refers to 0 or negative difference. The effect of an image characteristic is shown in the form of the log(odds ratio (OR)) representing the logarithms of the odds of improvement in the presence of the characteristic, compared to the odds of improvement occurring in the absence of that characteristic. The vertical black line refers to the point of equal odds. (Note: ***: p-value $\leq$ 0.001,**: p-value $\leq$ 0.01, *: p-value $\leq$ 0.05).}
  \label{fig:statistics}
\end{figure}

\begin{figure}[t!]
  \centering
  \begin{subfigure}{0.9\textwidth}
  \centering
    \includegraphics[width=\linewidth]{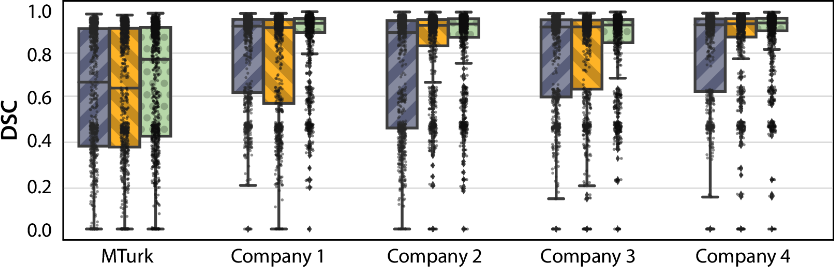}
    \caption{Annotation companies generate higher DSC scores than MTurk. \(0 \leq \text{DSC} \leq 1\).}
    \label{fig:provider_comparison_dsc}
  \end{subfigure}
  \hfill
  \begin{subfigure}{0.9\textwidth}
  \centering
    \includegraphics[width=\linewidth]{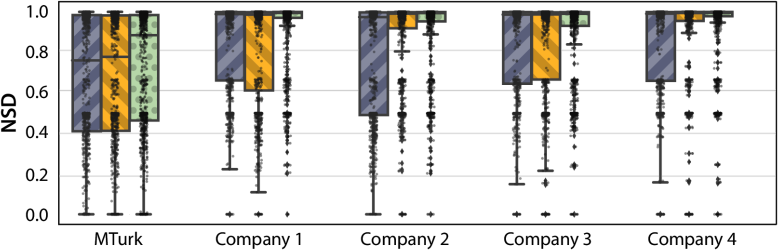}
    \caption{Annotation companies generate higher NSD scores than MTurk.\(0 \leq \text{NSD} \leq 1\).}
    \label{fig:provider_comparison_nsd}
  \end{subfigure}
  \hfill
  \begin{subfigure}{0.9\textwidth}
  \centering
    \includegraphics[width=\linewidth]{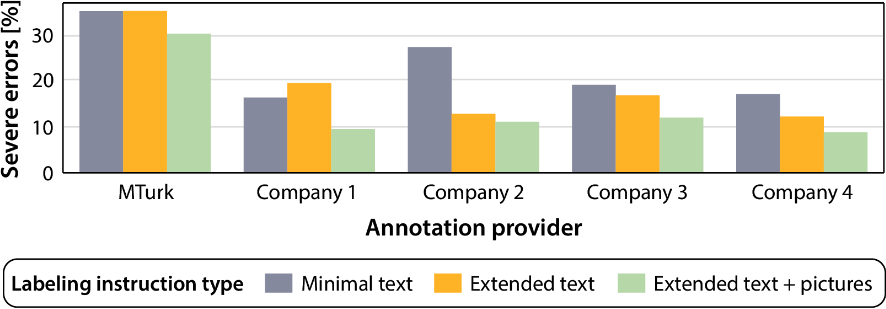}
    \caption{Annotation companies generate a lower number of severe errors.}
    \label{fig:provider_comparison_sev_errors}
  \end{subfigure}
  \caption{\textbf{Annotation companies generate higher quality annotations with their annotation pipeline than MTurk.} This is observed for (a) the DSC, (b) the NSD, and (c) number of severe errors (lower is better). Annotation companies include a QA step, in contrast to crowdsourcing on MTurk. (a) and (b) displayed as dots and box plots (band indicates the median, box indicates the first (25th percentile) and third (75th percentile) quartiles). }
  \label{fig:provider_comparison_general}
\end{figure}

As shown in Fig.~\ref{fig:statistics}\textcolor{red}{a}, QA significantly increases the odds of improving the annotation quality for images with generalizable image characteristics such as underexposed objects, intersection of objects, objects occluded by background in comparison to the absence of this characteristic (vertical black line). The only exceptions are motion artifacts present in the background and object(s) overexposure. 

In contrast to the general increase of images with generalizable image characteristics, none of the domain-specific image characteristics yield a significant change in the odds of improving the annotation quality over a regular image, see Fig.~\ref{fig:statistics}\textcolor{red}{b}. The only exception for domain-specific characteristics is object(s) covered by smoke, which could be interpreted as a form of object occlusion.  

\section{Discussion}
\label{sec:Discussion}
The data annotation market is estimated to reach a volume of more than USD 4.1 billion by the end of 2024~\cite{report_annotation_2024}. As the annotation process becomes increasingly automated and foundation models evolve, the market is undergoing drastic shifts. With lower entry barriers for annotation, more efficient data annotation tools, and a rising impact of QA, it is thought to result in an estimated USD 8.2 billion market by 2028~\cite{report_annotation_2030}. As a research community, we need to understand how to ensure high annotation quality throughout this evolution, as reliable reference annotations are an indispensable foundation for the benchmarking of image analysis algorithms, and thus their ultimate translation into real-world practice.

To our knowledge, this study is the first to quantitatively and critically examine the impact of internal QA conducted by annotation companies in the context of safety-critical applications. The most important insights can be summarized as follows:

\textbf{1. Annotation companies perform better both in terms of quantity and quality compared to the widely used platform MTurk} (Fig.~\ref{fig:budgets}). This implies that requesters working with crowdsourcing platforms should seriously consider annotation companies as an alternative. The organizational structure of these companies, which enables a single point of contact for requesters and eliminates spam annotations, appears to be a contributing factor to their efficiency. In light of poor working conditions being a major problem in annotation generation~\cite{miceli_data-production_2022,irani_cultural_2015}, annotation companies, which can more easily be inspected than distributed crowdsourcing platforms, are thought to offer better working contracts, enabling fairer working conditions for annotators. It should be noted that not every annotation company automatically ensures adequate working conditions. The annotation companies employed in this study underwent thorough scrutiny to guarantee compliance with good working condition guidelines. With annotators working at a shared location instead of in a distributed crowd, verifying annotators' working conditions has become substantially easier. We thus view our work as an initial step towards greater accountability on the part of annotation requesters. 

\textbf{2. In contrast to common belief, internal QA mechanisms do not necessarily improve annotation quality} (Fig.~\ref{fig:changes}). Furthermore, we observe no systematic QA quality improvement difference between annotation companies depending on their region of origin, size, or price, meaning more expensive annotation companies do not translate to better QA. However, improving labeling instructions instead of investing in QA can substantially boost the annotation performance (Fig.~\ref{fig:boxplot}). Consequently, in order to obtain high-quality annotations, annotation requesters should prioritize refining their labeling instructions, followed by implementing QA or even a different external QA process as a secondary step. Compared to QA, better labeling instructions can be generated with relatively few resources.

\textbf{3. The benefit of QA depends on the specific image characteristics} (Fig.~\ref{fig:statistics}). In fact, QA did improve the quality of images with certain challenging real-world image characteristics, see Fig.~\ref{fig:statistics}\textcolor{red}{a}. Consequently, it might be useful for requesters and annotation companies alike to allocate resources to identifying challenging image characteristics during a project, train classifiers to identify images containing these characteristics and focus the internal QA effort on these images.

One could contend that the recent rise of foundation models marginalizes the role of annotation providers. We argue, however, that it only changes their role. In this context, it is necessary to differentiate between the data quality demands for the training and testing of models. With a recent trend towards fine-tuning pre-trained models, smaller high-quality datasets can achieve similar performance with less computation~\cite{lee_platypus_2023,chen_alpagasus_2023}, highlighting the importance of the quality of unlabeled data in unsupervised training and its role in reducing carbon emissions~\cite{touvron_llama_2023}. In testing, especially for safety-critical applications such as autonomous driving and clinical medicine, ensuring high-quality annotated test data is crucial for accurate model evaluation and real-world applicability~\cite{sambasivan_everyone_2021}. Consequently, while training data annotation may lose relevance, test data annotation and its high quality demands are here to stay. Regarding foundation models, it should further be noted that annotations purely produced by foundation models introduce artifacts and biases~\cite{bommasani_opportunities_2022} and represent an increasing rising problem for MTurk, as demonstrated by Veselovsky et al.~\cite{veselovsky_artificial_2023}.

Our study is subject to several limitations which deserve further discussion. Comparing prices across providers is non-trivial as the costs to annotate a given data set depend on many factors including, among others, image quantity, type of annotations, location of annotation company and quality requirements. Our analysis was based on the actual costs incurred during our project. Confidentiality agreements with the software company which integrated our experiments into their annotation project pipeline prevent us from disclosing the identities of the annotation companies. Additionally, we took precautions to protect the identities of individual annotators and QA workers. This was particularly important given our findings that QA workers did not significantly enhance annotation quality, potentially exposing them to negative repercussions.

Another limitation could be seen in the fact that our data set only represents one safety-critical domain. While this is a valid concern, we would like to highlight that we chose a dataset which combined quality, complexity, and volume. Furthermore, regarding the breadth of annotation providers and the number of annotators involved, our study is likely the largest of its kind, thereby incurring substantial costs.

Finally, conducting multiple experiments with identical annotation companies and varying datasets could have potentially revealed our experimental design. Had the annotation companies been aware of our research, outcomes could potentially have been biased to appear more favorable, given companies’ inherent interest in demonstrating their services as reliable. We thus chose to conduct our experiments such that they could be seamlessly integrated into the software company’s annotation pipeline, reducing their likelihood of exposure.

\section{Conclusion and Future Work}
\label{sec:Conclusion}

In conclusion, our study underscores the pivotal role of enhancing the initial annotation quality, for instance via allocating resources to the improvement of labeling instructions, over QA in obtaining higher reference annotation quality in high-stakes AI~\cite{sambasivan_everyone_2021}. Future research should delve into the development and implementation of comprehensive training programs for annotators from annotation companies to further enhance initial annotation quality. Exploring the role of annotation companies in the context of increasingly automated annotation processes driven by foundation models presents another avenue for future research.

\section{Acknowledgements}
\label{sec:acknowledgements}
A part of this work was funded by M. Mengler and S. Funke - during their time at understand.ai -, and Helmholtz Imaging. T. Rädsch was supported by a scholarship from the Hanns Seidel
Foundation with funds from the Federal Ministry of Education and
Research Germany (BMBF). We thank M. Gelz, S. Strzysch and N. Kraft for their continuous support.

%
%
\bibliographystyle{splncs04}
\bibliography{main}

\begin{thebibliography}{10}
\providecommand{\url}[1]{\texttt{#1}}
\providecommand{\urlprefix}{URL }
\providecommand{\doi}[1]{https://doi.org/#1}

\bibitem{ali_objective_2020}
Ali, S., Zhou, F., Braden, B., Bailey, A., Yang, S., Cheng, G., Zhang, P., Li, X., Kayser, M., Soberanis-Mukul, R.D., Albarqouni, S., Wang, X., Wang, C., Watanabe, S., Oksuz, I., Ning, Q., Yang, S., Khan, M.A., Gao, X.W., Realdon, S., Loshchenov, M., Schnabel, J.A., East, J.E., Wagnieres, G., Loschenov, V.B., Grisan, E., Daul, C., Blondel, W., Rittscher, J.: An objective comparison of detection and segmentation algorithms for artefacts in clinical endoscopy. Scientific Reports  \textbf{10}(1), ~2748 (Feb 2020). \doi{10.1038/s41598-020-59413-5}, \url{https://www.nature.com/articles/s41598-020-59413-5}, number: 1 Publisher: Nature Publishing Group

\bibitem{bommasani_opportunities_2022}
Bommasani, R., Hudson, D.A., Adeli, E., Altman, R., Arora, S., von Arx, S., Bernstein, M.S., Bohg, J., Bosselut, A., Brunskill, E., Brynjolfsson, E., Buch, S., Card, D., Castellon, R., Chatterji, N., Chen, A., Creel, K., Davis, J.Q., Demszky, D., Donahue, C., Doumbouya, M., Durmus, E., Ermon, S., Etchemendy, J., Ethayarajh, K., Fei-Fei, L., Finn, C., Gale, T., Gillespie, L., Goel, K., Goodman, N., Grossman, S., Guha, N., Hashimoto, T., Henderson, P., Hewitt, J., Ho, D.E., Hong, J., Hsu, K., Huang, J., Icard, T., Jain, S., Jurafsky, D., Kalluri, P., Karamcheti, S., Keeling, G., Khani, F., Khattab, O., Koh, P.W., Krass, M., Krishna, R., Kuditipudi, R., Kumar, A., Ladhak, F., Lee, M., Lee, T., Leskovec, J., Levent, I., Li, X.L., Li, X., Ma, T., Malik, A., Manning, C.D., Mirchandani, S., Mitchell, E., Munyikwa, Z., Nair, S., Narayan, A., Narayanan, D., Newman, B., Nie, A., Niebles, J.C., Nilforoshan, H., Nyarko, J., Ogut, G., Orr, L., Papadimitriou, I., Park, J.S., Piech, C., Portelance, E., Potts, C.,
  Raghunathan, A., Reich, R., Ren, H., Rong, F., Roohani, Y., Ruiz, C., Ryan, J., Ré, C., Sadigh, D., Sagawa, S., Santhanam, K., Shih, A., Srinivasan, K., Tamkin, A., Taori, R., Thomas, A.W., Tramèr, F., Wang, R.E., Wang, W., Wu, B., Wu, J., Wu, Y., Xie, S.M., Yasunaga, M., You, J., Zaharia, M., Zhang, M., Zhang, T., Zhang, X., Zhang, Y., Zheng, L., Zhou, K., Liang, P.: On the {Opportunities} and {Risks} of {Foundation} {Models} (Jul 2022). \doi{10.48550/arXiv.2108.07258}, \url{http://arxiv.org/abs/2108.07258}

\bibitem{bragg_sprout_2018}
Bragg, J., Mausam, Weld, D.S.: Sprout: {Crowd}-{Powered} {Task} {Design} for {Crowdsourcing}. Proceedings of the ACM on User Interface Software and Technology  (2018). \doi{10.1145/3242587.3242598}

\bibitem{chang_revolt_2017}
Chang, J.C., Amershi, S., Kamar, E.: Revolt: {Collaborative} {Crowdsourcing} for {Labeling} {Machine} {Learning} {Datasets}. In: Proceedings of the 2017 {CHI} {Conference} on {Human} {Factors} in {Computing} {Systems}. pp. 2334--2346. {CHI} '17, Association for Computing Machinery, New York, NY, USA (May 2017). \doi{10.1145/3025453.3026044}

\bibitem{chen_alpagasus_2023}
Chen, L., Li, S., Yan, J., Wang, H., Gunaratna, K., Yadav, V., Tang, Z., Srinivasan, V., Zhou, T., Huang, H., Jin, H.: {AlpaGasus}: {Training} {A} {Better} {Alpaca} with {Fewer} {Data} (Nov 2023). \doi{10.48550/arXiv.2307.08701}, \url{http://arxiv.org/abs/2307.08701}

\bibitem{cheplygina_early_2016}
Cheplygina, V., Perez-Rovira, A., Kuo, W., Tiddens, H.A.W.M., de~Bruijne, M.: Early {Experiences} with {Crowdsourcing} {Airway} {Annotations} in {Chest} {CT}. In: Carneiro, G., Mateus, D., Peter, L., Bradley, A., Tavares, J.M.R.S., Belagiannis, V., Papa, J.P., Nascimento, J.C., Loog, M., Lu, Z., Cardoso, J.S., Cornebise, J. (eds.) Deep {Learning} and {Data} {Labeling} for {Medical} {Applications}. pp. 209--218. Lecture {Notes} in {Computer} {Science}, Springer International Publishing, Cham (2016). \doi{10.1007/978-3-319-46976-8_22}

\bibitem{chmielewski_mturk_2020}
Chmielewski, M., Kucker, S.C.: An {MTurk} {Crisis}? {Shifts} in {Data} {Quality} and the {Impact} on {Study} {Results}. Social Psychological and Personality Science  \textbf{11}(4),  464--473 (May 2020). \doi{10.1177/1948550619875149}, publisher: SAGE Publications Inc

\bibitem{report_annotation_2024}
Cognilytica: Data engineering, preparation, and labeling for ai 2020 (2020), https://www.cognilytica.com/2020/01/31/data-preparation-labeling-for-ai-2020/

\bibitem{crequit_mapping_2018}
Créquit, P., Mansouri, G., Benchoufi, M., Vivot, A., Ravaud, P.: Mapping of {Crowdsourcing} in {Health}: {Systematic} {Review}. Journal of Medical Internet Research  \textbf{20}(5), ~187 (2018). \doi{10.2196/jmir.9330}, \url{https://www.jmir.org/2018/5/e187/}

\bibitem{daniel_quality_2019}
Daniel, F., Kucherbaev, P., Cappiello, C., Benatallah, B., Allahbakhsh, M.: Quality {Control} in {Crowdsourcing}: {A} {Survey} of {Quality} {Attributes}, {Assessment} {Techniques} and {Assurance} {Actions}. ACM Computing Surveys  \textbf{51},  1--40 (Jan 2019). \doi{10.1145/3148148}

\bibitem{dice_measures_1945}
Dice, L.R.: Measures of the amount of ecologic association between species. Ecology  \textbf{26}(3),  297--302 (1945)

\bibitem{dow_shepherding_2012}
Dow, S., Kulkarni, A., Klemmer, S., Hartmann, B.: Shepherding the crowd yields better work. In: Proceedings of the {ACM} 2012 conference on {Computer} {Supported} {Cooperative} {Work}. pp. 1013--1022. {CSCW} '12, Association for Computing Machinery, New York, NY, USA (Feb 2012). \doi{10.1145/2145204.2145355}

\bibitem{duggan_gamified_2023}
Duggan, N.M., Jin, M., Mendicuti, M.A.D., Hallisey, S., Bernier, D., Selame, L.A., Asgari-Targhi, A., Fischetti, C.E., Lucassen, R., Samir, A.E., Duhaime+, E., Kapur, T., Goldsmith, A.J.: Gamified {Crowdsourcing} as a {Novel} {Approach} to {Lung} {Ultrasound} {Dataset} {Labeling} (Jun 2023). \doi{10.48550/arXiv.2306.06773}, \url{http://arxiv.org/abs/2306.06773}

\bibitem{gaikwad2017prototype}
Gaikwad, S., Chhibber, N., Sehgal, V., Ballav, A., Mullings, C., Nasser, A., Richmond-Fuller, A., Gilbee, A., Gamage, D., Whiting, M., et~al.: Prototype tasks: improving crowdsourcing results through rapid, iterative task design  (2017). \doi{10.48550/arXiv.1707.05645}, \url{https://arxiv.org/abs/1707.05645}

\bibitem{halevy_unreasonable_2009}
Halevy, A., Norvig, P., Pereira, F.: The unreasonable effectiveness of data. IEEE intelligent systems  \textbf{24}(2),  8--12 (2009). \doi{10.1109/MIS.2009.36}, iSBN: 1541-1672 Publisher: IEEE

\bibitem{han2023neglected}
Han, D., Choe, J., Chun, S., Chung, J.J.Y., Chang, M., Yun, S., Song, J.Y., Oh, S.J.: Neglected free lunch--learning image classifiers using annotation byproducts  (2023). \doi{10.48550/arXiv.2303.17595}, \url{https://arxiv.org/abs/2303.17595}

\bibitem{heim_large-scale_2018}
Heim, E., Roß, T., Seitel, A., März, K., Stieltjes, B., Eisenmann, M., Lebert, J., Metzger, J., Sommer, G., Sauter, A.W.: Large-scale medical image annotation with crowd-powered algorithms. Journal of Medical Imaging  \textbf{5}(3),  034002 (2018). \doi{10.1117/1.JMI.5.3.034002}

\bibitem{hettiachchi_challenge_2021}
Hettiachchi, D., Schaekermann, M., McKinney, T.J., Lease, M.: The {Challenge} of {Variable} {Effort} {Crowdsourcing} and {How} {Visible} {Gold} {Can} {Help}. Proceedings of the ACM on Human-Computer Interaction  \textbf{5},  332:1--332:26 (Oct 2021). \doi{10.1145/3476073}

\bibitem{hutson_quality_2019}
Hutson, M., Kanzheleva, O., Taggart, C., Campana, B.J.L., Duong, Q.: Quality {Control} {Challenges} in {Crowdsourcing} {Medical} {Labeling}  (2019)

\bibitem{idrissi_imagenet-x_2022}
Idrissi, B.Y., Bouchacourt, D., Balestriero, R., Evtimov, I., Hazirbas, C., Ballas, N., Vincent, P., Drozdzal, M., Lopez-Paz, D., Ibrahim, M.: {ImageNet}-{X}: {Understanding} {Model} {Mistakes} with {Factor} of {Variation} {Annotations} (Nov 2022). \doi{10.48550/arXiv.2211.01866}, \url{http://arxiv.org/abs/2211.01866}

\bibitem{ipeirotis_quality_2010}
Ipeirotis, P.G., Provost, F., Wang, J.: Quality management on {Amazon} {Mechanical} {Turk}. In: Proceedings of the {ACM} {SIGKDD} {Workshop} on {Human} {Computation}. pp. 64--67. ACM, Washington DC (Jul 2010). \doi{10.1145/1837885.1837906}, \url{https://dl.acm.org/doi/10.1145/1837885.1837906}

\bibitem{irani_cultural_2015}
Irani, L.: The cultural work of microwork. New Media \& Society  \textbf{17}(5),  720--739 (May 2015). \doi{10.1177/1461444813511926}, publisher: SAGE Publications

\bibitem{irshad_crowdsourcing_2014}
Irshad, H., Montaser-Kouhsari, L., Waltz, G., Bucur, O., Nowak, J., Dong, F., Knoblauch, N., Beck, A.: Crowdsourcing image annotation for nucleus detection and segmentation in computational pathology: evaluating experts, automated methods, and the crowd. In: Biocomputing 2015. pp. 294--305. WORLD SCIENTIFIC, Kohala Coast, Hawaii, USA (Nov 2014). \doi{10.1142/9789814644730_0029}, \url{http://www.worldscientific.com/doi/abs/10.1142/9789814644730_0029}

\bibitem{joskowicz_inter-observer_2019}
Joskowicz, L., Cohen, D., Caplan, N., Sosna, J.: Inter-observer variability of manual contour delineation of structures in {CT}. European Radiology  \textbf{29}(3),  1391--1399 (Mar 2019). \doi{10.1007/s00330-018-5695-5}, \url{http://link.springer.com/10.1007/s00330-018-5695-5}

\bibitem{kentley_agreement_2023}
Kentley, J., Weber, J., Liopyris, K., Braun, R.P., Marghoob, A.A., Quigley, E.A., Nelson, K., Prentice, K., Duhaime, E., Halpern, A.C., Rotemberg, V.: Agreement {Between} {Experts} and an {Untrained} {Crowd} for {Identifying} {Dermoscopic} {Features} {Using} a {Gamified} {App}: {Reader} {Feasibility} {Study}. JMIR Medical Informatics  \textbf{11},  e38412 (Jan 2023). \doi{10.2196/38412}, \url{https://medinform.jmir.org/2023/1/e38412}

\bibitem{kohli_medical_2017}
Kohli, M.D., Summers, R.M., Geis, J.R.: Medical image data and datasets in the era of machine learning—whitepaper from the 2016 {C}-{MIMI} meeting dataset session. Journal of digital imaging  \textbf{30}(4),  392--399 (2017). \doi{10.1007/s10278-017-9976-3}, publisher: Springer

\bibitem{le_ensuring_2010}
Le, J., Edmonds, A., Hester, V., Biewald, L.: Ensuring quality in crowdsourced search relevance evaluation: The effects of training question distribution. In: SIGIR 2010 workshop on crowdsourcing for search evaluation. vol.~2126, pp. 22--32 (2010)

\bibitem{le_ludec_problem_2023}
Le~Ludec, C., Cornet, M., Casilli, A.A.: The problem with annotation. {Human} labour and outsourcing between {France} and {Madagascar}. Big Data \& Society  \textbf{10}(2),  20539517231188723 (Jul 2023). \doi{10.1177/20539517231188723}, \url{http://journals.sagepub.com/doi/10.1177/20539517231188723}

\bibitem{lee_platypus_2023}
Lee, A.N., Hunter, C.J., Ruiz, N.: Platypus: {Quick}, {Cheap}, and {Powerful} {Refinement} of {LLMs} (Aug 2023). \doi{10.48550/arXiv.2308.07317}, \url{http://arxiv.org/abs/2308.07317}

\bibitem{litman_relationship_2015}
Litman, L., Robinson, J., Rosenzweig, C.: The relationship between motivation, monetary compensation, and data quality among {US}- and {India}-based workers on {Mechanical} {Turk}. Behavior Research Methods  \textbf{47}(2),  519--528 (Jun 2015). \doi{10.3758/s13428-014-0483-x}, \url{http://link.springer.com/10.3758/s13428-014-0483-x}

\bibitem{lu_rethinking_2023}
Lu, X., Ratcliffe, D., Kao, T.T., Tikhonov, A., Litchfield, L., Rodger, C., Wang, K.: Rethinking {Quality} {Assurance} for {Crowdsourced} {Multi}-{ROI} {Image} {Segmentation}. Proceedings of the AAAI Conference on Human Computation and Crowdsourcing  \textbf{11}(1),  103--114 (Nov 2023). \doi{10.1609/hcomp.v11i1.27552}, \url{https://ojs.aaai.org/index.php/HCOMP/article/view/27552}, number: 1

\bibitem{maier-hein_can_2014}
Maier-Hein, L., Mersmann, S., Kondermann, D., Bodenstedt, S., Sanchez, A., Stock, C., Kenngott, H.G., Eisenmann, M., Speidel, S.: Can masses of non-experts train highly accurate image classifiers? In: International conference on medical image computing and computer-assisted intervention. pp. 438--445. Springer (2014). \doi{10.1007/978-3-319-10470-6_55}

\bibitem{maier-hein_metrics_2024}
Maier-Hein, L., Reinke, A., Godau, P., Tizabi, M.D., Buettner, F., Christodoulou, E., Glocker, B., Isensee, F., Kleesiek, J., Kozubek, M., et~al.: Metrics reloaded: recommendations for image analysis validation. Nature Methods  \textbf{21},  195--212 (2024). \doi{10.1038/s41592-023-02151-z}, \url{https://www.nature.com/articles/s41592-023-02151-z}

\bibitem{maier-hein_heidelberg_2021}
Maier-Hein, L., Wagner, M., Ross, T., Reinke, A., Bodenstedt, S., Full, P.M., Hempe, H., Mindroc-Filimon, D., Scholz, P., Tran, T.N., Bruno, P., Kisilenko, A., Müller, B., Davitashvili, T., Capek, M., Tizabi, M.D., Eisenmann, M., Adler, T.J., Gröhl, J., Schellenberg, M., Seidlitz, S., Lai, T.Y.E., Pekdemir, B., Roethlingshoefer, V., Both, F., Bittel, S., Mengler, M., Mündermann, L., Apitz, M., Kopp-Schneider, A., Speidel, S., Nickel, F., Probst, P., Kenngott, H.G., Müller-Stich, B.P.: Heidelberg colorectal data set for surgical data science in the sensor operating room. Scientific Data  \textbf{8}(1), ~101 (Apr 2021). \doi{10.1038/s41597-021-00882-2}, \url{https://www.nature.com/articles/s41597-021-00882-2}, number: 1 Publisher: Nature Publishing Group

\bibitem{mcculloch_generalized_2004}
McCulloch, C.E., Searle, S.R.: Generalized, linear, and mixed models. John Wiley \& Sons (2004). \doi{10.1002/0471722073}, \url{https://books.google.com/books?hl=en&lr=&id=bWDPukohugQC&oi=fnd&pg=PR6&dq=Generalized,+Linear,+and+Mixed+Models.&ots=sSy3HM7CGo&sig=Ss04EHKG_xkcZZCYWSEJl6paHos}

\bibitem{miceli_data-production_2022}
Miceli, M., Posada, J.: The {Data}-{Production} {Dispositif}. Proceedings of the ACM on Human-Computer Interaction  \textbf{6}(CSCW2),  1--37 (Nov 2022). \doi{10.1145/3555561}, \url{https://dl.acm.org/doi/10.1145/3555561}

\bibitem{michaelis_benchmarking_2020}
Michaelis, C., Mitzkus, B., Geirhos, R., Rusak, E., Bringmann, O., Ecker, A.S., Bethge, M., Brendel, W.: Benchmarking {Robustness} in {Object} {Detection}: {Autonomous} {Driving} when {Winter} is {Coming} (Mar 2020). \doi{10.48550/arXiv.1907.07484}, \url{http://arxiv.org/abs/1907.07484}

\bibitem{petrovic_crowdsourcing_2020}
Petrović, N., Moyà-Alcover, G., Varona, J., Jaume-i Capó, A.: Crowdsourcing human-based computation for medical image analysis: {A} systematic literature review. Health Informatics Journal  \textbf{26}(4),  2446--2469 (Dec 2020). \doi{10.1177/1460458220907435}, \url{http://journals.sagepub.com/doi/10.1177/1460458220907435}

\bibitem{reinke_understanding_2024}
Reinke, A., Tizabi, M.D., Baumgartner, M., Eisenmann, M., Heckmann-Nötzel, D., Kavur, A.E., Rädsch, T., Sudre, C.H., Acion, L., Antonelli, M.: Understanding metric-related pitfalls in image analysis validation. Nature methods pp. 1--13 (2024). \doi{doi.org/10.1038/s41592-023-02150-0}, \url{https://www.nature.com/articles/s41592-023-02150-0}, publisher: Nature Publishing Group

\bibitem{report_annotation_2030}
Research, G.V.: Data {Collection} {And} {Labeling} {Market} {Size} \& {Share} {Report} 2030 (2022), \url{https://www.grandviewresearch.com/industry-analysis/data-collection-labeling-market}

\bibitem{ros_how_2021}
Roß, T., Bruno, P., Reinke, A., Wiesenfarth, M., Koeppel, L., Full, P.M., Pekdemir, B., Godau, P., Trofimova, D., Isensee, F.: How can we learn (more) from challenges? {A} statistical approach to driving future algorithm development. arXiv preprint arXiv:2106.09302  (2021). \doi{10.48550/arXiv.2106.09302}, \url{https://arxiv.org/abs/2106.09302}

\bibitem{radsch_labelling_2023}
Rädsch, T., Reinke, A., Weru, V., Tizabi, M.D., Schreck, N., Kavur, A.E., Pekdemir, B., Roß, T., Kopp-Schneider, A., Maier-Hein, L.: Labelling instructions matter in biomedical image analysis. Nature Machine Intelligence  \textbf{5}(3),  273--283 (Mar 2023). \doi{10.1038/s42256-023-00625-5}, \url{https://www.nature.com/articles/s42256-023-00625-5}, number: 3 Publisher: Nature Publishing Group

\bibitem{sambasivan_everyone_2021}
Sambasivan, N., Kapania, S., Highfill, H., Akrong, D., Paritosh, P., Aroyo, L.M.: “{Everyone} wants to do the model work, not the data work”: {Data} {Cascades} in {High}-{Stakes} {AI}. In: Proceedings of the 2021 {CHI} {Conference} on {Human} {Factors} in {Computing} {Systems}. pp. 1--15. ACM, Yokohama Japan (May 2021). \doi{10.1145/3411764.3445518}, \url{https://dl.acm.org/doi/10.1145/3411764.3445518}

\bibitem{graham_planetary_2022}
Schmidt, F.A.: The {Planetary} {Stacking} {Order} of {Multilayered} {Crowd}-{AI} {Systems}. In: Graham, M., Ferrari, F. (eds.) Digital {Work} in the {Planetary} {Market}, pp. 137--156. The MIT Press (Jun 2022). \doi{10.7551/mitpress/13835.003.0012}, \url{https://direct.mit.edu/books/book/5319/chapter/3800159/The-Planetary-Stacking-Order-of-Multilayered-Crowd}

\bibitem{touvron_llama_2023}
Touvron, H., Lavril, T., Izacard, G., Martinet, X., Lachaux, M.A., Lacroix, T., Rozière, B., Goyal, N., Hambro, E., Azhar, F., Rodriguez, A., Joulin, A., Grave, E., Lample, G.: {LLaMA}: {Open} and {Efficient} {Foundation} {Language} {Models} (Feb 2023). \doi{10.48550/arXiv.2302.13971}, \url{http://arxiv.org/abs/2302.13971}

\bibitem{veselovsky_artificial_2023}
Veselovsky, V., Ribeiro, M.H., West, R.: Artificial {Artificial} {Artificial} {Intelligence}: {Crowd} {Workers} {Widely} {Use} {Large} {Language} {Models} for {Text} {Production} {Tasks} (Jun 2023). \doi{10.48550/arXiv.2306.07899}, \url{http://arxiv.org/abs/2306.07899}

\bibitem{zheng_improving_2023}
Wang, K., Hill, M., Knowles-Barley, S., Tikhonov, A., Litchfield, L., Bare, J.C.: Improving {Segmentation} of {Breast} {Arterial} {Calcifications} from {Digital} {Mammography}: {Good} {Annotation} is {All} {You} {Need}. In: Zheng, Y., Keleş, H.Y., Koniusz, P. (eds.) Computer {Vision} – {ACCV} 2022 {Workshops}, vol. 13848, pp. 134--150. Springer Nature Switzerland, Cham (2023). \doi{10.1007/978-3-031-27066-6_10}, \url{https://link.springer.com/10.1007/978-3-031-27066-6_10}, series Title: Lecture Notes in Computer Science

\bibitem{yu_difficulty-aware_2020}
Yu, S., Zhou, H.Y., Ma, K., Bian, C., Chu, C., Liu, H., Zheng, Y.: Difficulty-{Aware} {Glaucoma} {Classification} with {Multi}-rater {Consensus} {Modeling}. In: Martel, A.L., Abolmaesumi, P., Stoyanov, D., Mateus, D., Zuluaga, M.A., Zhou, S.K., Racoceanu, D., Joskowicz, L. (eds.) Medical {Image} {Computing} and {Computer} {Assisted} {Intervention} – {MICCAI} 2020. pp. 741--750. Lecture {Notes} in {Computer} {Science}, Springer International Publishing, Cham (2020). \doi{10.1007/978-3-030-59710-8_72}

\bibitem{orting_survey_2020}
Ørting, S.N., Doyle, A., Hilten, A.v., Hirth, M., Inel, O., Madan, C.R., Mavridis, P., Spiers, H., Cheplygina, V.: A {Survey} of {Crowdsourcing} in {Medical} {Image} {Analysis}. Human Computation  \textbf{7},  1--26 (Dec 2020). \doi{10.15346/hc.v7i1.1}, \url{https://hcjournal.org/index.php/jhc/article/view/111}

\end{thebibliography}

\clearpage
\setcounter{page}{1}
\renewcommand{\thesection}{\Alph{section}}
\setcounter{section}{0}

\renewcommand{\figurename}{Supplementary Fig.}
\setcounter{figure}{0}

\colorbox{BurntOrange}{Disclaimer:
The Supplementary contains bloody surgery images.} 

\clearpage

\section{Real-world image characteristics examples}
\label{sec:app_A_image_characteristics}
The 57,636 metadata annotations represent real-world image characteristics and were created by three engineers with the help of some crowdworkers. Examples are displayed in Supplementary Fig.~\ref{fig:app_a}. The characteristics can be generalizable (Supplementary Fig.~\ref{fig:app_a}\textcolor{red}{a}) or domain-specific (Supplementary Fig.~\ref{fig:app_a}\textcolor{red}{b}), and occur in the background, on the objects or as an overlay. Generalizable characteristics include: motion artifacts in the background, other objects present in the background, object occluded by background, intersection of objects, motion artifacts on the object, object(s) overexposure, object(s) underexposure. Domain-specific characteristics include: background covered by blood, background covered by reflections, background covered by smoke, image overlaid with text, caption at the bottom of the image, black box present in the image, trocar present in the image, object covered by blood, object(s) covered by reflections, object(s) covered by smoke.

\begin{figure}
  \centering
  \includegraphics[width=0.75\textwidth]{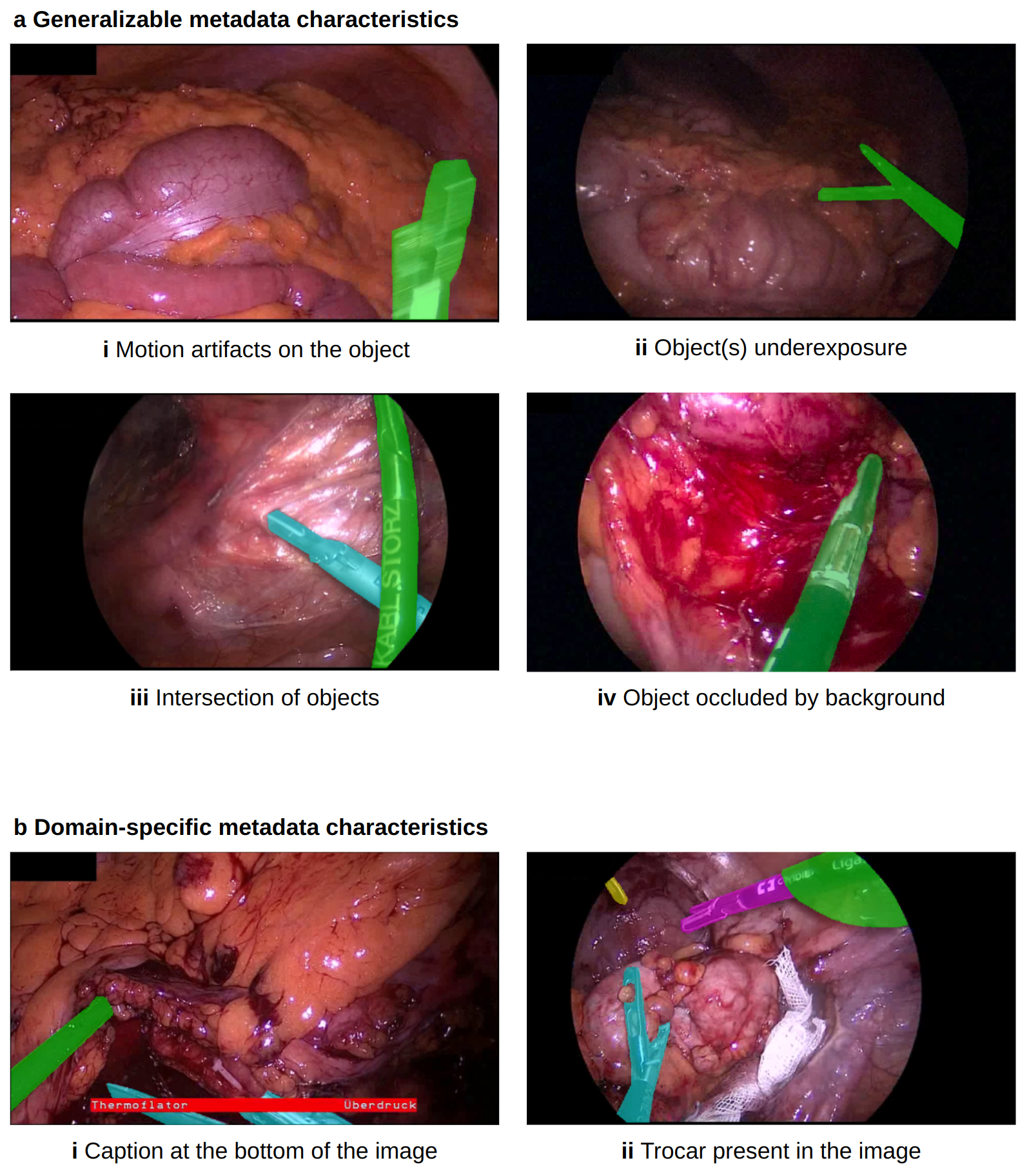}
  \caption{\textbf{Real-world image characteristics examples.} During recording, a broad range of (a) generalizable and (b) domain-specific image characteristics occur and can influence humans and models alike.}
  \label{fig:app_a}
\end{figure}

\includepdf[pages=1-,pagecommand=\section{Labeling instruction types}
\label{sec:app_B_li}\subsection{Minimal text labeling instruction\label{sec_sub:li_min}},nup = 2x4, offset=0 0, frame=true, scale=0.7]{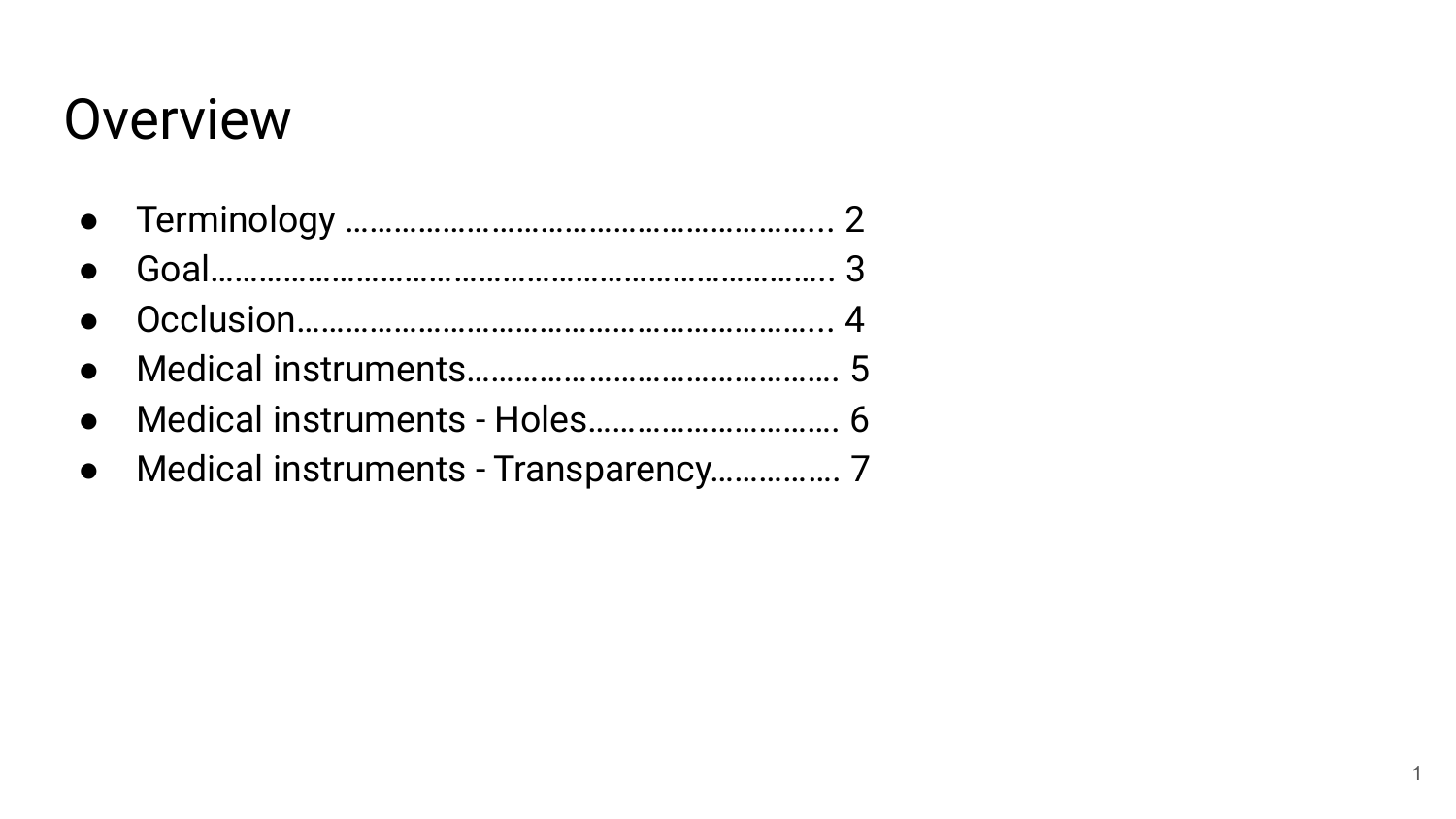}
\clearpage

\includepdf[pages=1-8,pagecommand=\subsection{Extended text labeling instruction}\label{sec_sub:li_ext},nup = 2x4, offset=0 0, frame=true, scale=0.7]{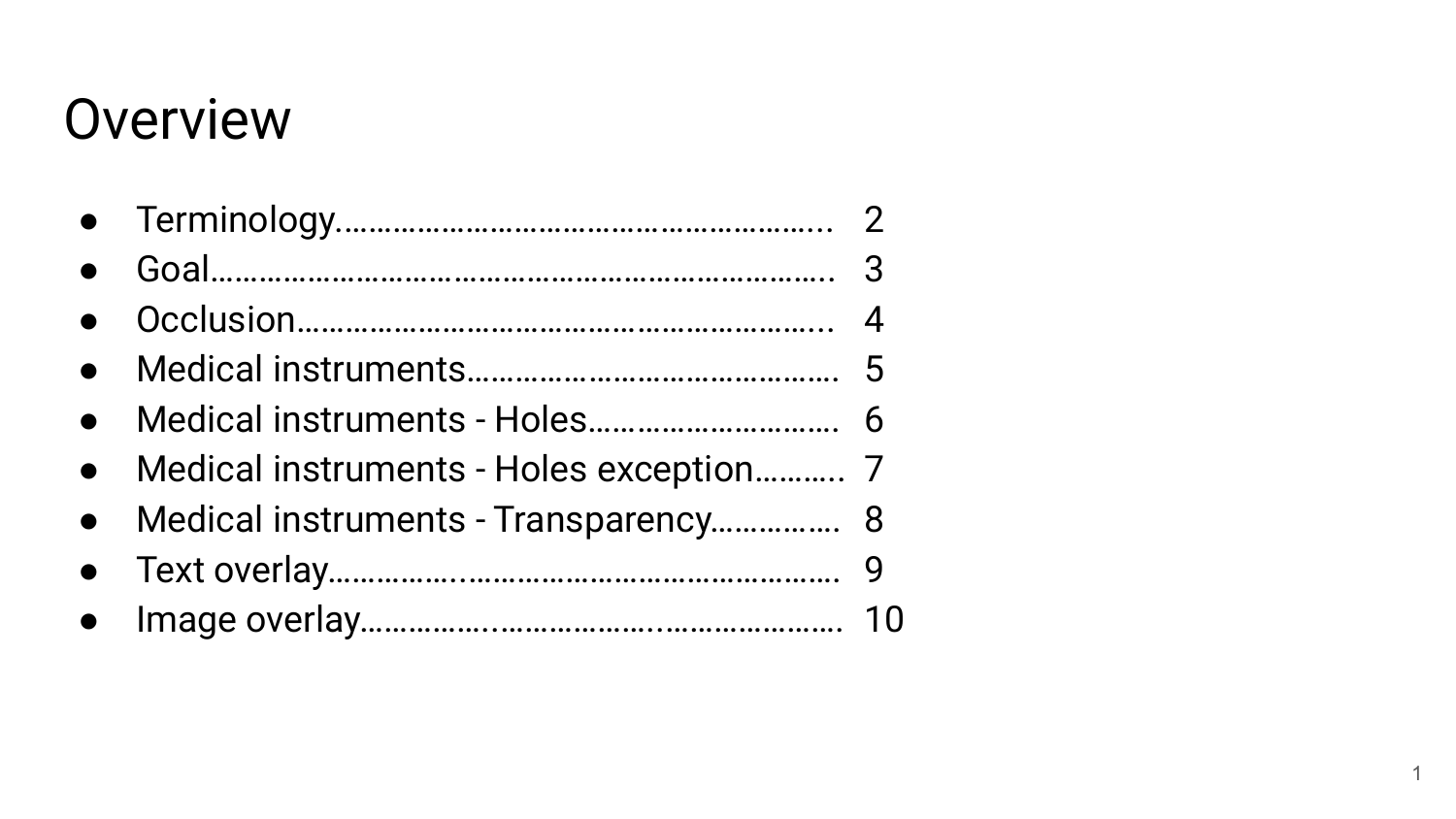}
\clearpage
\includepdf[pages=9-,pagecommand={},nup = 2x4, offset=0 0, frame=true, scale=0.7]{appendix_files/instruction_02_extended_text.pdf}
\clearpage

\includepdf[pages=1-8,pagecommand=\subsection{Extended text including pictures labeling instruction}
\label{sec_sub:li_pic},nup = 2x4, offset=0 0, frame=true, scale=0.7]{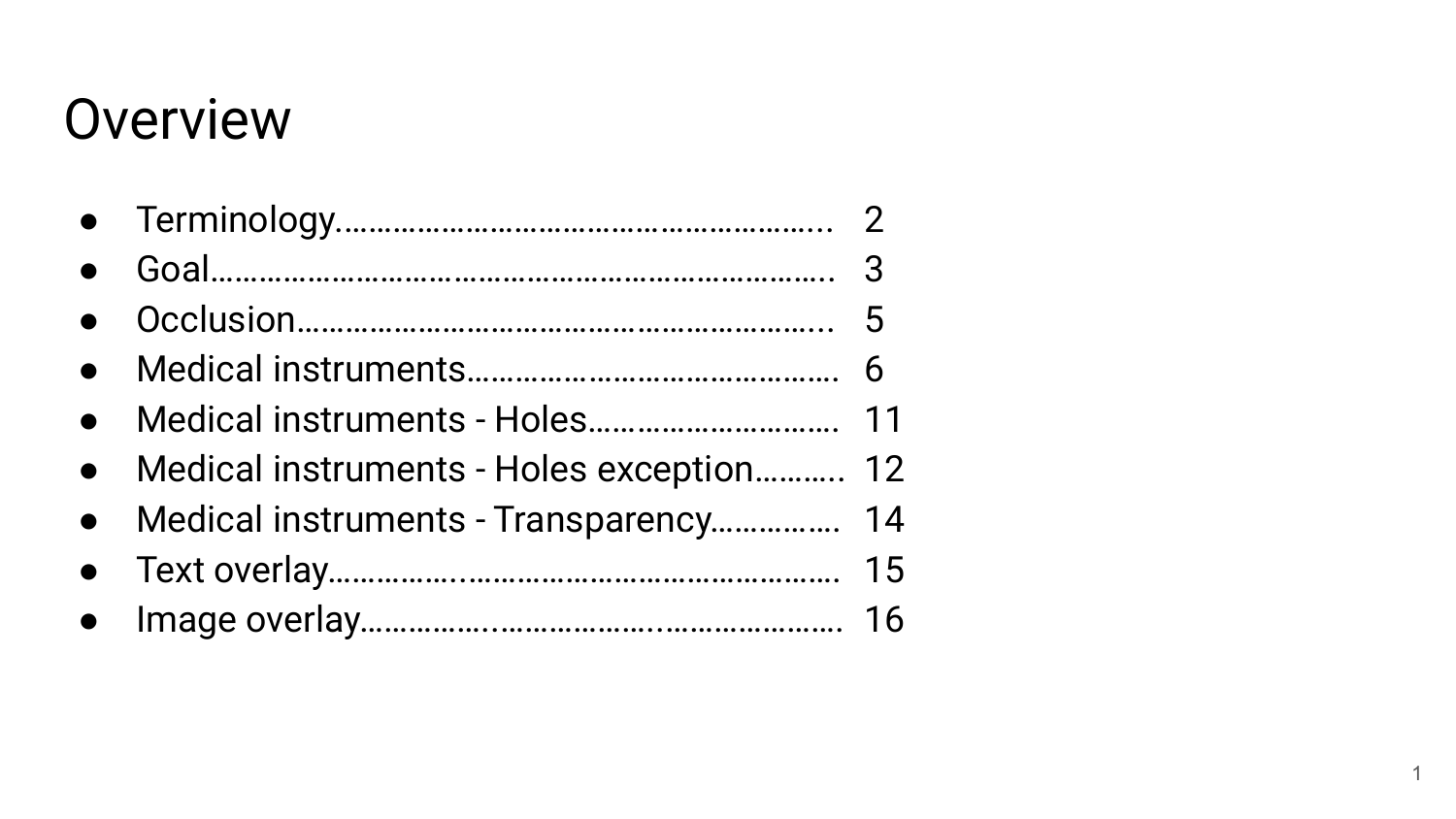}
\clearpage
\includepdf[pages=9-,pagecommand={},nup = 2x4, offset=0 0, frame=true, scale=0.7]{appendix_files/instruction_03_picture_version.pdf}

\section{Additional method details}
\label{sec:app_C_li}

\textbf{Analysis of computer vision conference exhibitors}\\
We included the major computer vision conferences from 2022 and 2023, namely: ECCV 2022, CVPR 2022, ICCV 2023, and CVPR 2023. The list of exhibitors was extracted from the respective conference websites\footnote{Accessed 2024-03-04: https://hallerickson.ungerboeck.com/prod/app85.cshtml?aat=\\\{wRAm7m6yFc6aFoOwv9ggPME6PgM3DwEgvwcFfgyN3YY\%3d;\\Cr9Jl1p0qL33RLclLrfrlBKRmhROPXGkDwcQAVv7zHM\%3D;\\Zva\%2bsPD7WeTWWGsGWvarzixed2IJHWExCd84dKNS6Yg\%3d\};\\https://iccv2023.thecvf.com/exhibitor.list.and.floor.plan-365000-5-44-50.php}. We define an annotation provider as an entity that provides annotations performed by human workers. Consequently, each exhibitor's website was analyzed to determine whether the exhibitor provides human-generated annotations as a core component of their offering. Companies offering solely synthetic data were not included. In the last two years, around 20\% of the conference exhibitors were annotation providers, as illustrated in Fig.~\ref{fig:workflows_and_exhibitors}. As the majority of exhibitors were offering specific tech products, we excluded exhibitor booths from our analysis that were not clearly related to a specific offering (e.g., the booth for the ECCV coffee station or the Springer Publishing Group). This resulted in three excluded exhibitors for ECCV 2022 (total before exclusion: 43), seven excluded exhibitors for CVPR 2022 (total before exclusion: 104), three excluded exhibitors for ICCV 2023 (total before exclusion: 47), and eight excluded exhibitors for CVPR 2023 (total before exclusion: 112).
\\\\
\textbf{Annotation tools}\\
MTurk operates on a self-service basis. In this model, the individual or organization requesting annotations is responsible for supplying all aspects, including training for annotators and the annotation tools, whereas the crowdworkers are engaged as independent contractors. In contrast, annotation companies provide a managed service for handling annotators, who can work on any annotation tool that the requester gives access to. Since the tool used for the annotation companies was not available for MTurk, we used a different annotation tool for MTurk. Both annotation tools were developed with best design practices to ensure the production of high-quality annotations. Furthermore, the tools were tested internally, until a sufficient user experience and annotation result were achieved. As intermediate annotations without QA are typically not visible to the requester, the software company added automated backend exports after the annotation step to enable this study. Professional annotators are aware that their work is internally verified by QA workers. None of the annotators participating had previous experience with the annotation tools they were using.
\\\\
\textbf{Logistic mixed model}\\
The logistic mixed model was implemented in the lme4 package in R. The obtained estimates of the covariates are on the log-odds scale and were exponentiated to obtain the odds ratio for each covariate. Software: R version 4.0.2 (package lme4 version 1.1.33).

\begin{equation}
\text{I} \sim (1 | \text{Image\_name}) + (1 | \text{Company} / \text{Annotator}) + \sum_{i=1}^{7} G_i  + \sum_{i=1}^{10} D_i
\end{equation}

where:
\begin{itemize}
    \item $\text{Improvement (I)}$ is the response variable.
    \item $(1 | \text{Image\_name})$ represents a random intercept for each \text{Image\_name}.
    \item $(1 | \text{Company} / \text{Annotator})$ represents a  random intercept, with \text{Annotator} nested within \text{Company}.
    \item $\sum_{i=1}^{7} G_i$ represents the generalizable fixed effects, with $G_i$ being the $i^{th}$ analyzed generalizable covariate.
    \item $\sum_{i=1}^{10} D_i$ represents the domain-specific fixed effects, with $D_i$ being the $i^{th}$ analyzed domain-specific covariate.
\end{itemize}

List of covariates:
\begin{itemize}
  \item Generalizable covariates:
  \begin{itemize}
  \item Motion artifacts in the background
  \item  Other objects present in the background
  \item Object occluded by background
  \item Intersection of objects
  \item Motion artifacts on the object
  \item Object(s) overexposure
  \item Object(s) underexposure
  \end{itemize}
  \item  Domain-specific covariates:
  \begin{itemize}
  \item Background covered by blood
  \item Background covered by reflections
  \item Background covered by smoke
  \item Image overlaid with text
  \item Caption at the bottom of the image
  \item Black box present in the image
  \item Trocar present in the image
  \item Object covered by blood
  \item Object(s) covered by reflections
  \item Object(s) covered by smoke
  \end{itemize}
\end{itemize}

Code will be published with acceptance. 
\\\\
\textbf{Annotation provider analysis}\\
The 57,636 metadata annotations represent different real-world imaging characteristics metadata. Utilizing these meta-annotations, we sorted the images of the dataset into nine categories that reflect the complexity of annotation, which guided the selection process for subsequent image analysis:
\\
(1)
Simple category: The image does not contain any artifacts on the instruments.
\\
(2)
Chaos category: The image contains at least three different artifacts on the instruments, and those containing a greater number of instruments are preferred.
\\
(3)
Trocar category: The image contains at least one trocar.
\\
(4)
Intersection category: The image contains at least two medical instruments intersecting.
\\
(5)
Motion blur category: The image contains at least one medical instrument exhibiting the motion blur artifact.
\\
(6)
Underexposure category: The image contains at least one medical instrument that is underexposed.
\\
(7)
Text overlay category: The image contains text overlay that obstructs the view.
\\
(8)
Image overlay category: The image contains an image overlay that obstructs the view of the image.
\\
(9)
Random category: Images are randomly selected from the remaining images in the test set.

For the comparison between the different annotation provider types, we chose 234 distinct frames in line with the specified categories. We manually picked 15 distinct images from each of the first eight categories, which made up about half of the total images. The exception was category 8, which only had 11 unique images fit its criteria. The remainder of the images were randomly selected from category 9. Each frame was labeled four times following each set of labeling instructions and by each annotation provider, accumulating 60 annotations for each frame. Each image annotation of the companies was checked by QA workers (culminating in 25,272 labeled images for the provider comparison). While most annotation companies prioritize sustained work packages with longer completion times, MTurk focuses on microtasks that are completed rapidly and in parallel. Thus, every professional annotator completed annotations for a total of 72 images and each MTurk crowdworker annotated four images, to accurately replicate the parallelization seen in crowdsourcing.

Each annotated image from all annotation providers for both analyses was checked by an engineer in our team to identify spam annotations. The annotation companies generated no spam annotations, whereas MTurk annotated images contained a proportion of spam annotations of around 20\%. Examples of spam annotations are displayed in Supplementary Fig.~\ref{fig:app_spam_annotations}.
\\\\
\textbf{Real-world image condition analysis}\\
For the impact analysis of real-world image characteristics on QA, we focused solely on the extended text including pictures labeling instructions and the annotation companies. All 4,050 unique frames of the HeiCo dataset were annotated two times by each annotation company, accumulating eight annotations for each frame.
For this analysis we utilized all 57,636 metadata annotations.

\clearpage
\section{Spam annotation examples}
\label{sec:app_D_li}
\begin{figure}
  \centering
  \includegraphics[width=0.9\textwidth]{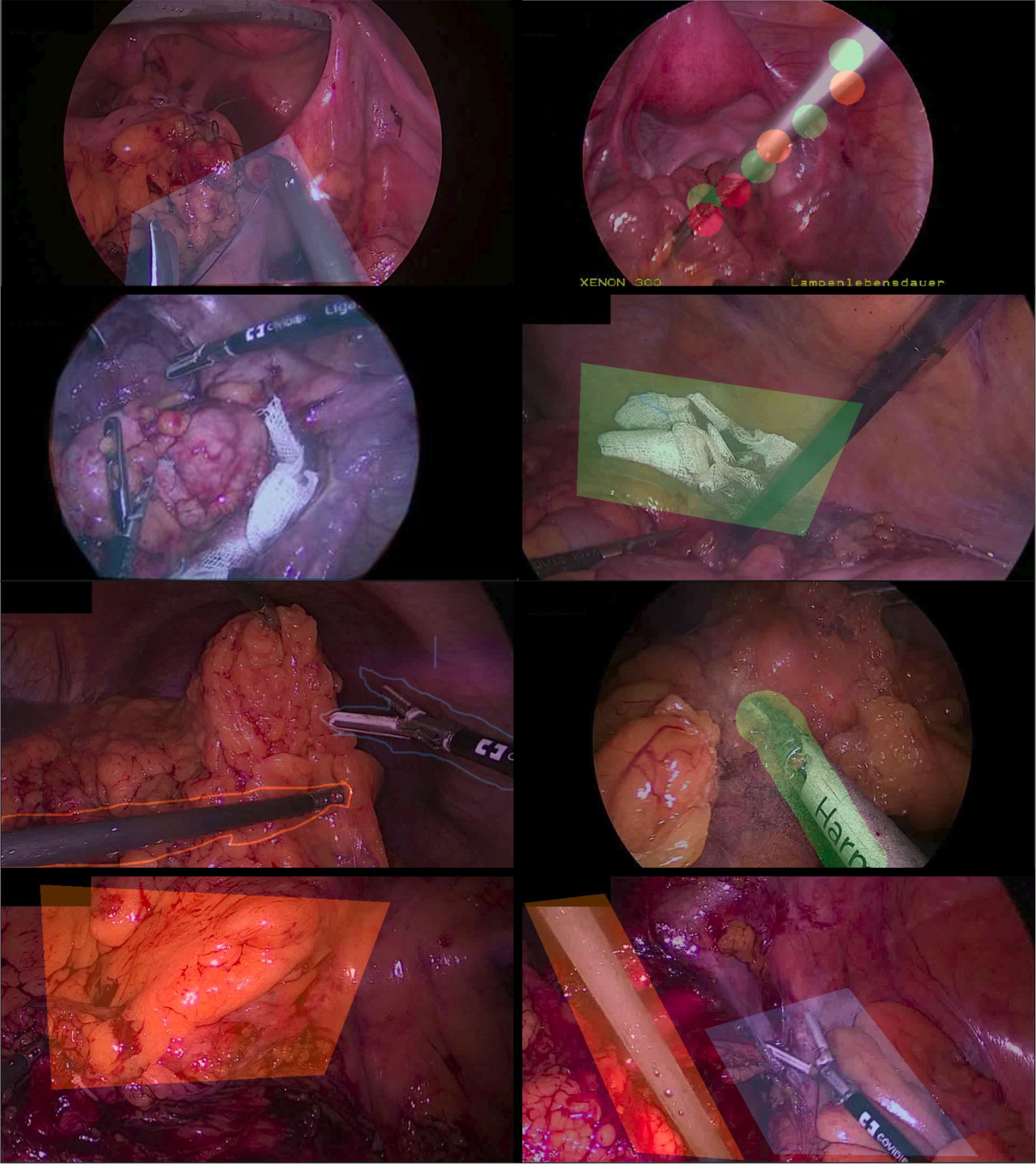}
  \caption{\textbf{Only MTurk crowdworkers generated spam annotations.} The annotation companies generated no spam annotations, whereas around 20\% of the collected MTurk annotations were spam. The spam annotations vary greatly in their style and seem to display different strategies to game the MTurk annotation tasks.}
  \label{fig:app_spam_annotations}
\end{figure}

\clearpage
\section{Impact of annotation companies' QA on NSD scores}
\label{sec:app_E_li}

        \begin{table}
            \caption{QA of annotation companies only provides marginal improvements of the Normalized Surface Distance (NSD) scores, if any.}
            \label{tab:nsd_d}
            \centering
            \scriptsize
            \begin{tabular}{ll|cc|cc|cc|}
 & & \multicolumn{2}{c}{25th Percentile}& \multicolumn{2}{c}{Median}& \multicolumn{2}{c}{75th Percentile}\\
                 Company&  Stage&  Annotate&  QA  &  Annotate&  QA  &  Annotate& QA  \\
                 \midrule
                 Company 1&  1) Minimal text &  0.65&  0.66&  0.99&  0.99&  1& 1\\
                 Company 1&  2) Extended text &  0.62&  0.61&  0.99&  0.99&  1& 1\\
                 Company 1&  3) Extended text + pictures &  0.97&  0.97&  0.99&  0.99&  1& 1\\
                                  \midrule
                 Company 2&  1) Minimal text &  0.61&  0.49&  0.98&  0.97&  1& 1\\
                 Company 2&  2) Extended text &  0.66&  0.92&  0.99&  0.99&  1& 1\\
                 Company 2&  3) Extended text + pictures &  0.93&  0.95&  0.99&  0.99&  1& 1\\
                                  \midrule
 Company 3& 1) Minimal text & 0.63& 0.65& 0.99& 0.99& 1&1\\
 Company 3& 2) Extended text & 0.5& 0.67& 0.98& 0.99& 1&1\\
     
 Company 3& 3) Extended text + pictures & 0.82& 0.93& 0.99& 0.99& 1&1\\
  \midrule
 Company 4& 1) Minimal text & 0.66& 0.66& 0.99& 0.99& 1&1\\            Company 4&  2) Extended text &  0.96&  0.96&  0.99&  0.99&  1& 1\\
                 Company 4&  3) Extended text + pictures &  0.98&  0.98&  1&  1&  1& 1\\
            \end{tabular}

        \end{table}


\section{Impact of annotation companies' QA on severe errors}
\label{sec:app_E_li}

        \begin{table}
            \caption{QA of annotation companies only provides marginal reductions of severe errors. These depend highly on the company and type of labeling instruction.}
            \label{tab:sev_error_e}
            \centering
            \scriptsize

            \begin{tabular}{ll|cc|c|}
                 Company&  Stage&  Annotate&  QA&  Difference\\
                 \midrule
                 Company 1&  1) Minimal text &  0.18&  0.17&  -0.01\\
                 Company 1&  2) Extended text &  0.2&  0.2&  0\\
                 Company 1&  3) Extended text + pictures &  0.1&  0.1&  0\\
                                  \midrule
                 Company 2&  1) Minimal text &  0.22&  0.28&  0.06\\
                 Company 2&  2) Extended text &  0.19&  0.13&  -0.06\\
                 Company 2&  3) Extended text + pictures &  0.14&  0.11&  -0.02\\
                                  \midrule
 Company 3& 1) Minimal text & 0.21& 0.19& -0.01\\
 Company 3& 2) Extended text & 0.25& 0.17& -0.08\\
     
 Company 3& 3) Extended text + pictures & 0.16& 0.12& -0.04\\
  \midrule
 Company 4& 1) Minimal text & 0.18& 0.17& 0\\            Company 4&  2) Extended text &  0.12&  0.12&  0\\
                 Company 4&  3) Extended text + pictures &  0.09&  0.09&  0\\
            \end{tabular}
        \end{table}


\clearpage
\section{Modification distribution by QA workers}
The number of modified images depended highly on the annotation company, as displayed in Supplementary Subfig.~\ref{fig:app_f}\textcolor{red}{a}. 
Nevertheless, a substantial proportion of modified images did not inherently imply superior annotation quality (Supplementary Subfig.~\ref{fig:app_f}\textcolor{red}{b}). For instance, Company 1 on average made modifications to every third annotated image, while Company 4 only made modifications to every 10th image. However, it is evident that Company 4's resulting annotations clearly outperformed those of Company 1 when employing the extended text and extended text including pictures labeling instructions. The combination of a low modification rate and high annotation quality achieved by Company 4 could suggest a higher level of annotator expertise.

Overall, the annotations conducted with the extended text including pictures labeling instructions outperform all other experiments and QA efforts. This is a clear indication that further research on internal QA processes is needed.

\begin{figure}
  \centering
  \includegraphics[width=0.9\textwidth]{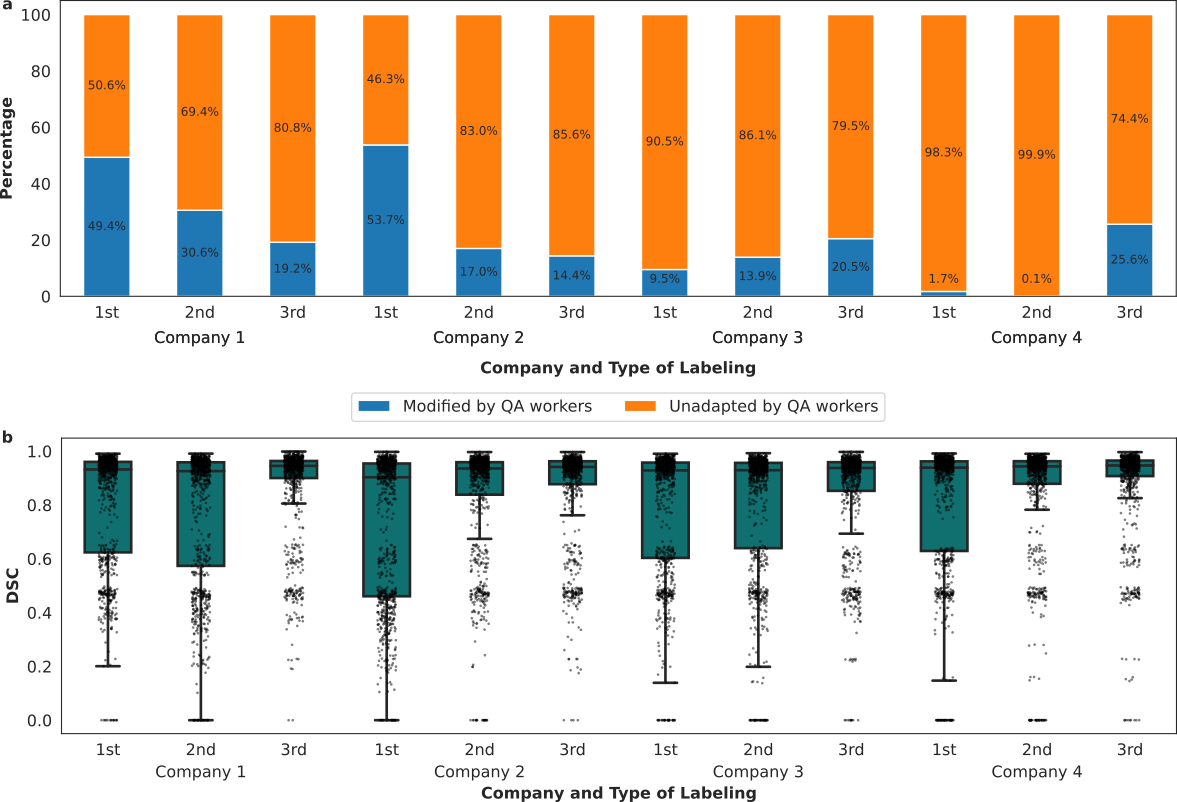}
  \caption{\textbf{The percentage of modifications by QA workers differed greatly between the companies. However, this did not necessarily result in higher annotation quality, when compared to other annotation companies.} 
  \textbf{a}, The percentages of modified and unadapted images by QA workers. An image counts as modified if the responsible QA worker adapted the received instance segmentation mask from the annotator. \textbf{b}, The DSC was aggregated for each resulting annotated image and is displayed aggregated for each pair of annotation company and labeling instruction as a dots- and boxplot (the band indicates the median, the box indicates the first and third quartiles and the whiskers indicate $ \pm 1.5 \times$ interquartile range), the DSC maximum is 1 and the minimum is 0 for each image.}
  \label{fig:app_f}
\end{figure}

\clearpage

\end{document}